\documentclass[letterpaper]{article} 
\usepackage{aaai2027}  
\usepackage[hyphens]{url}  
\usepackage{graphicx} 
\usepackage{natbib}  
\usepackage{caption} 
\usepackage{algorithm}
\usepackage{algorithmic}
\usepackage[utf8]{inputenc} 
\usepackage[T1]{fontenc}    

\usepackage{url}            
\usepackage{booktabs}       
\usepackage{amsfonts}       
\usepackage{nicefrac}       
\usepackage{microtype}      
\usepackage{xcolor}         
\usepackage{times}
\usepackage{graphicx}

\usepackage{multicol}
\usepackage{multirow}
\usepackage{amsmath} 
\usepackage{amssymb} 
\usepackage{array}
\usepackage{marvosym}
\usepackage{caption}
\usepackage{amsmath}
\usepackage{tabularx}

\newcommand{\eqcell}[2]{%
  \refstepcounter{equation}%
  \label{#2}%
  \hbox to \hsize{%
    {\small$\displaystyle #1$}%
    \hfil
    \textup{(\theequation)}%
  }%
}

\usepackage{newfloat}
\usepackage{listings}
\DeclareCaptionStyle{ruled}{labelfont=normalfont,labelsep=colon,strut=off} 
\floatstyle{ruled}
\newfloat{listing}{tb}{lst}{}
\floatname{listing}{Listing}

\usepackage{booktabs}

\title{\textbf{SeedPolicy}: Horizon Scaling via
\textbf{Se}lf-\textbf{E}volving \textbf{D}iffusion Policy\\
for Robot Manipulation}

\author{
    Youqiang Gui\textsuperscript{\rm 1,*,$\dagger$},
    Yuxuan Zhou\textsuperscript{\rm 3,*,$\dagger$},
    Shen Cheng\textsuperscript{\rm 2},
    Xinyang Yuan\textsuperscript{\rm 1},\\
    Haoqiang Fan\textsuperscript{\rm 2},
    Peng Cheng\textsuperscript{\rm 1,$\ddagger$},
    Shuaicheng Liu\textsuperscript{\rm 4,$\ddagger$}
}

\affiliations{
    \mbox{\textsuperscript{\rm 1}Sichuan University}
    \quad
    \mbox{\textsuperscript{\rm 2}Dexmal Inc.}
    \quad
    \mbox{\textsuperscript{\rm 3}Independent Researcher}
    \quad
    \mbox{\textsuperscript{\rm 4}UESTC}
    \\
    \textsuperscript{*}Equal contribution
    \quad
    \textsuperscript{$\dagger$}Work done during internship at Dexmal Inc.
    \quad
    \textsuperscript{$\ddagger$}Corresponding authors
}

\begin{document}

\maketitle

\begin{abstract}
  Imitation Learning (IL) enables robots to acquire manipulation skills from expert demonstrations. Diffusion Policy (DP) models multi-modal expert behaviors but degrades when naively increasing stacked observation horizons, limiting long-horizon manipulation.
We propose Self-Evolving Gated Attention (SEGA), a temporal module that maintains a time-evolving latent state via gated attention, enabling efficient recurrent updates that accumulate long-term context into a compact latent representation while filtering irrelevant temporal information. Integrating SEGA into DP yields \textbf{Se}lf-\textbf{E}volving \textbf{D}iffusion \textbf{Policy} (SeedPolicy), which resolves the temporal modeling bottleneck and extends the effective temporal horizon with moderate overhead.
On the RoboTwin 2.0 benchmark with 50 manipulation tasks, SeedPolicy outperforms DP and other IL baselines. Averaged across both CNN and Transformer backbones, SeedPolicy achieves 36.8\% relative improvement in clean settings and 169\% relative improvement in randomized challenging settings over the DP. Compared to vision–language–action models such as RDT with 1.2B parameters, SeedPolicy achieves stronger performance in the clean setting with one to two orders of magnitude fewer parameters, demonstrating strong efficiency. These results establish SeedPolicy as a state-of-the-art imitation learning method for long-horizon robotic manipulation. 
\end{abstract}

\section{Introduction}
Imitation Learning (IL) has emerged as a dominant paradigm in embodied AI, enabling robots to learn versatile manipulation skills directly from expert demonstrations~\cite{3,16,17,1,20,21,23, 27}. Transformer-based methods, such as ACT~\cite{1}, mitigate jitter and enhance trajectory smoothness by predicting action chunks instead of single-step actions. Building on this, Diffusion Policy~\cite{3} introduced diffusion models to robotic control, explicitly capturing the multi-modal distribution of human behaviors and achieving unprecedented stability and precision in complex tasks.

\begin{figure*}[t]
    \centering
    \includegraphics[width= 0.8\linewidth]{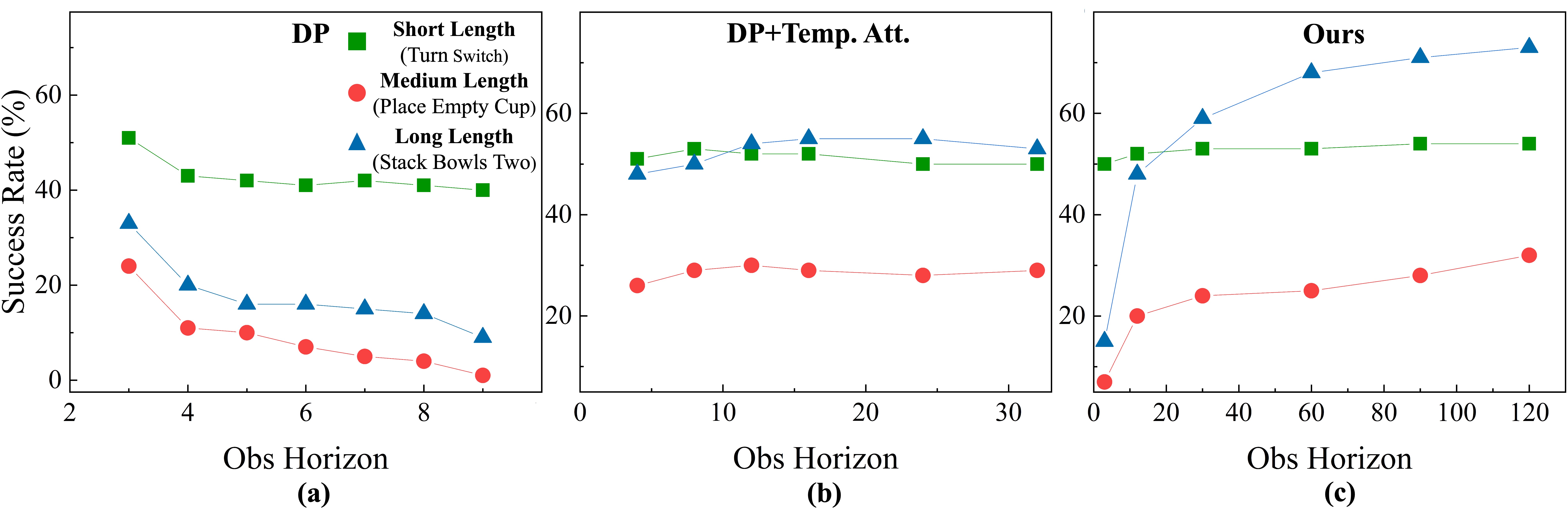}
    \caption{\textbf{Horizon scaling analysis.} (a) DP shows a counter-intuitive performance drop as the observation horizon grows, indicating that naive frame stacking fails to scale. (b) Temporal attention can initially improve performance, but scaling it to longer horizons incurs quadratic cost and diminishing returns. (c) In contrast, our approach improves task success by recurrently extending the effective temporal receptive field without increasing the per-step observation window.}
    \label{fig:1}
    \vspace{-10pt}
\end{figure*}

Despite its prominence, we identify a critical limitation in the modeling capabilities of standard Diffusion Policies. As shown in Fig.~\ref{fig:1} (a), naively increasing the stacked observation horizon of the baseline policy paradoxically degrades performance. While the  authors briefly acknowledged this counter-intuitive phenomenon in the appendix~\cite{3}, the underlying causes were not explored or resolved. In this work, we reveal that this degradation stems from a fundamental limitation: simply treating the observations as a larger stack of image frames~\cite{3,4,5} fails to capture complex temporal dependencies, an issue that becomes more pronounced as the number of frames grows.

To bridge this gap, we add temporal self-attention over stacked observation features to enable explicit cross-timestep interaction. As shown in Fig.~\ref{fig:1} (b), this straightforward yet effective solution shows that explicit temporal modeling can better exploit stacked histories than simple frame stacking. Empirically, it provides a clear advantage in capturing long-term dependencies, validating our initial analysis.

However, the computational cost of attention layers increases quadratically with the observation horizon, and extending temporal attention to longer horizons often leads to diminishing returns, making it difficult to scale effectively. To address this, we propose a recurrent-style update mechanism, where our framework maintains a time-evolving latent state that continuously encodes the historical context. By condensing continuous information into a latent representation, this design significantly reduces computational overhead while effectively capturing long-term dependencies.

Furthermore, in dynamic robotic manipulation, valuable information is often temporally sparse; not every observation contributes meaningfully to the task. Visual disturbances such as irrelevant background shifts or occlusions can introduce noise, and indiscriminately integrating these frames risks polluting historical context. Motivated by the benefits of increased sparsity in attention scores achieved through gating mechanisms~\cite{19}, we incorporate Self-Evolving Gate (SEG) into our temporal modeling framework. Unlike conventional gating strategies, SEG utilizes the cross-attention logits as a regulatory mechanism. Specifically, SEG dynamically suppresses noisy or irrelevant signals and modulates the state evolution process, ensuring that only semantically relevant information is preserved. We denote this integrated mechanism as Self-Evolving Gated Attention (SEGA).

When combined with Diffusion Policy (DP), we term it \textbf{Se}lf-\textbf{E}volving \textbf{D}iffusion \textbf{Policy} (SeedPolicy). To the best of our knowledge, this is the first method to effectively resolve the temporal modeling bottleneck in Diffusion Policy. As shown in Fig.~\ref{fig:1} (c), our approach reverses the baseline's trend: our performance consistently improves as the observation horizon scales up.

Our main contributions are summarized as follows: \textbf{(1)} We propose Self-Evolving Gated Attention (SEGA), a temporal module that synergizes attention with a dynamic gating mechanism to maintain a compact evolving latent state, capturing long-term dependencies while filtering irrelevant temporal disturbances.
\textbf{(2)} We demonstrate effective horizon scaling for diffusion-based control, reversing the performance degradation seen in prior diffusion policies and consistently turning a longer effective temporal receptive field into measurable performance gains.
\textbf{(3)} We introduce SeedPolicy, achieving state-of-the-art among imitation learning methods on the RoboTwin 2.0 benchmark: an average of 36.8\% improvement in clean settings and 169\% improvement in randomized, challenging scenarios over Diffusion Policy. Compared with the 1.2B-parameter RDT, SeedPolicy achieves stronger clean-setting performance with up to 36× fewer parameters.

\section{Method}

\begin{figure*}[t]
    \centering
    \includegraphics[width= 0.9\linewidth]{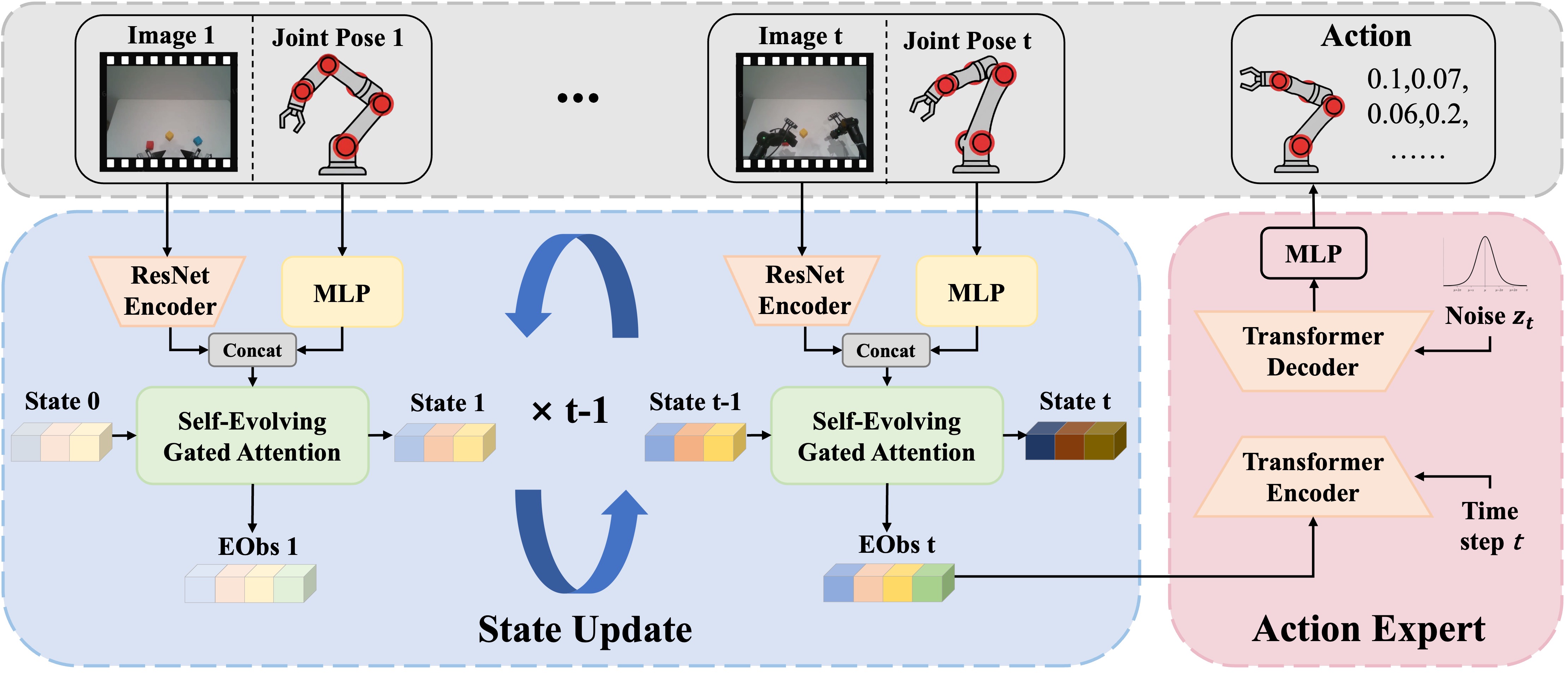}
    \caption{\textbf{Overview of the SeedPolicy framework.} 
The system takes current RGB images and joint poses as input, encoding them via a ResNet Encoder and MLP. 
The core \textbf{Self-Evolving Gated Attention (SEGA)} module (blue box) recursively updates a time-evolving latent state ($State \ t$) to capture long-term spatiotemporal dependencies while generating enhanced observation features ($EObs_t$). 
These context-rich features are then fed into the Action Expert, a transformer-based diffusion model, to predict a sequence of future actions.}
    \label{fig:framework}
    \vspace{-8pt}
\end{figure*}
  
\subsection{Overview of SeedPolicy}

SeedPolicy is an end-to-end framework for robotic manipulation, as illustrated in Fig.~\ref{fig:framework}. 
First, the current RGB image $I_t$ and joint pose $P_t$ are encoded by a ResNet Encoder into observation features $O_t \in \mathbb{R}^{N_o \times D}$, where $N_o$ denotes the number of observation feature vectors and $D$ represents the feature dimension.  
To capture long-term spatiotemporal dependencies and enforce temporal sparsity, we propose the \textbf{Self-Evolving Gated Attention (SEGA)} module. 
This module maintains a time-evolving latent state $S_{t-1} \in \mathbb{R}^{N_s \times D}$, where $N_s$ denotes the context size of the latent state encoding historical information.
The per-step observation window remains fixed, while the effective temporal receptive field is extended by accumulating and preserving long-term information in the evolving latent state.
Specifically, SEGA facilitates bidirectional interaction: it simultaneously updates the state with new observation to produce the updated state $S_t$ (Eq. (2)), and utilizes historical context to generate enhanced observation features $EObs_t$ (Eq. (3)).
Finally, $EObs_t$ are fed into a Diffusion Action Expert to generate a sequence of $N$ future 14-DoF actions $A_t$.

The overall process at time step $t$ can be formalized as:
\begingroup
\setlength{\tabcolsep}{0pt}

\noindent
\begin{tabularx}{\linewidth}{
  @{}
  X
  @{\hspace{1.2em}}
  X
  @{}
}

\eqcell{
  O_t = \operatorname{Encoder}(I_t, P_t)
}{eq:encoder}
&
\eqcell{
  S_t = \operatorname{Update}(S_{t-1}, O_t)
}{eq:update}
\\[6pt]

\eqcell{
  \mathit{EObs}_t =
  \operatorname{Retrieve}(O_t, S_{t-1})
}{eq:retrieve}
&
\eqcell{
  A_t =
  \operatorname{Diffusion}(\mathit{EObs}_t)
}{eq:diffusion}

\end{tabularx}
\endgroup

\begin{figure*} 
    \centering
    \includegraphics[width= 0.9\linewidth]{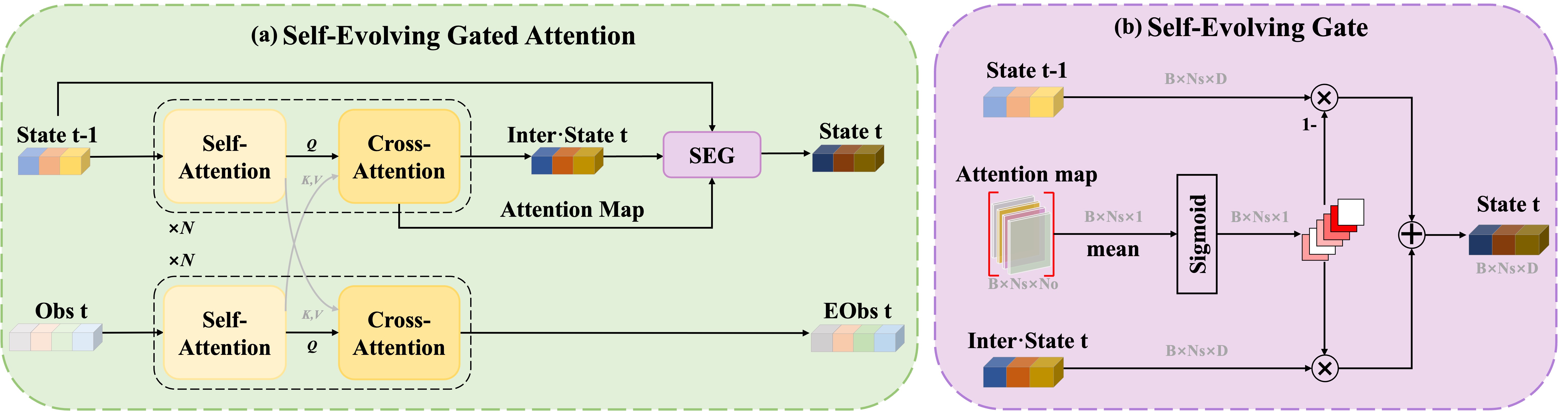}
    \caption{(a) SEGA employs a dual-stream design: the \textbf{State Update} stream (top) evolves the latent state ($State_{t-1}$) by integrating new observations, while the \textbf{State Retrieval} stream (bottom) utilizes historical context to generate enhanced observation features ($EObs_t$). 
(b) The \textbf{Self-Evolving Gate (SEG)} dynamically computes a gating signal directly from the cross-attention logits. It selectively fuses the intermediate evolved state ($\text{Inter} \cdot S_{t}$) with the previous state, ensuring only semantically relevant information is preserved while filtering out noise.}
    \label{fig:sega}
    \vspace{-8pt}
\end{figure*}

\subsection{Self-Evolving Gated Attention}
As illustrated in Fig.~\ref{fig:sega} (a), SEGA employs a parallel dual-stream Transformer design that facilitates continuous interaction between the historical latent state $S_{t-1}$ and the current observation $O_t$. Specifically, the module coordinates two parallel processes:
(1) \textbf{State Update} (Upper stream), which evolves the latent state by integrating new sensory information and regulating the update intensity via gate mechanism to produce the final state $S_t$.
(2) \textbf{State Retrieval} (Lower stream), which utilizes the historical context to enrich the current sensory input, retrieving relevant temporal cues to generate the enhanced observation features $EObs_t$.

\noindent \textbf{State Update.} 
The upper stream of Fig.~\ref{fig:sega}~(a) is dedicated to maintaining the time-evolving latent state. This process involves two critical steps: extracting new information and regulating its integration. 
First, both the historical state $S_{t-1}$ and the current observation $O_t$ undergo Multi-Head Self-Attention (MSA) to extract internal contextual features:

\begingroup

\setlength{\tabcolsep}{0pt}

\noindent
\begin{tabularx}{\linewidth}{
  @{}
  X
  @{\hspace{1.2em}}
  X
  @{}
}

\eqcell{
  S_{t-1}' =
  S_{t-1} + \operatorname{MSA}(S_{t-1})
}{eq:state_msa}
&
\eqcell{
  O_t' =
  O_t + \operatorname{MSA}(O_t)
}{eq:obs_msa}

\end{tabularx}
\endgroup

Subsequently, the new state features $S'_{t-1}$ act as the \textit{Query}, extracting relevant semantic information from the observation features $O'_t$ (acting as \textit{Key} and \textit{Value}). This produces the intermediate state $\text{Inter} \cdot S_{t}$ and the attention logits $\mathcal{A}$:
\begin{equation}
   \text{Inter} \cdot S_{t}, \mathcal{A} = \text{CA}(S'_{t-1}, O'_t, O'_t)
\end{equation}
where $\mathcal{A} = \{A^{(l, h)} \mid l \in [1, L], h \in [1, H]\}$ denote the pre-softmax logits from all layers and heads.

Indiscriminately integrating every observation risks polluting the state with visual disturbances (e.g., background shifts or distracting objects). To enforce temporal sparsity, we incorporate the Self-Evolving Gate (SEG) at the end of this stream (Fig.~\ref{fig:sega} (b)).
SEG interprets the raw attention scores as ``relevance signals'' to adaptively regulate the update. We compute a global relevance score $R$ and derive the update gate $G_t \in \mathbb{R}^{N_s \times 1}$ as follows:
\begin{equation}
    G_t = \sigma(R)
\end{equation}
where
\begin{equation}
    R = \frac{1}{L \cdot H \cdot N_o} \sum_{l=1}^{L} \sum_{h=1}^{H} \sum_{j=1}^{N_o} A^{(l, h)}_{:, j}
\end{equation}

The final updated state $S_t$ is a gated fusion, ensuring only semantically relevant information is preserved:
\begin{equation}
    S_t = G_t \odot \text{Inter} \cdot S_{t} + (1 - G_t) \odot S_{t-1}
\end{equation}

Our mechanism does not introduce inference bottlenecks. In robot control, observations naturally arrive sequentially. At step $t$, the previous state $S_{t-1}$ is already computed and cached. SeedPolicy seamlessly updates this state using the new observation $O_t$.

\noindent \textbf{State Retrieval.} 
Simultaneously, the lower stream of Fig.~\ref{fig:sega} (a) leverages the accumulated historical context to enrich the current sensory input. 
In contrast to the update path, the interaction roles are reversed here: the processed observation features $O'_t$ serve as the \textit{Query}, actively seeking relevant temporal clues from the historical state features $S'_{t-1}$, which act as both \textit{Key} and \textit{Value}. 
This mechanism is crucial for bridging the horizon gap, allowing the model to recover information lost due to long-term dependencies.
The resulting enhanced observation $EObs_t$ is computed via:
\begin{equation}
    EObs_t = \text{CA}(O'_t, S'_{t-1}, S'_{t-1})
\end{equation}
These context-enriched features are subsequently forwarded to the Action Expert, providing a robust perceptual basis for precise action prediction.

\section{Experiments} 
\label{sec:experiments}

We design experiments to answer the following questions: \textbf{Q1:} Does our design improve policy performance in robot manipulation? \textbf{Q2:} What are the capabilities of SeedPolicy? \textbf{Q3:} How does SEGA compare with existing memory mechanisms? \textbf{Q4:} Which design choices matter for SeedPolicy?

\subsection{Setup}
\subsubsection{Simulation Benchmark}
We evaluate SeedPolicy on RoboTwin 2.0~\cite{15}, RMBench~\cite{48}, and MimicGen~\cite{50}. 
On RoboTwin 2.0, we evaluate 50 manipulation tasks with ALOHA-AgileX. Each policy is trained with 50 expert demonstrations for 600 epochs and tested with 100 rollouts per task. We report the mean success rate over three independent trials under both Easy (\textit{demo\_clean}; train/test clean) and Hard (\textit{demo\_randomized}; clean-trained, randomized-test) settings. 
For RMBench~\cite{48} and MimicGen~\cite{50}, we follow original settings and evaluation protocols~\cite{48, 49}. Detailed experimental analyses are provided in Appendix Sec.~4.

\subsubsection{Real Robot Benchmark}
We evaluate SeedPolicy on the Dexmal DOS W1 robot using a fixed Intel RealSense D435 RGB camera. For each task, we collect 50 demonstrations, train for 600 epochs, and conduct two evaluation trials of 50 rollouts each. We assess robustness to \textit{state ambiguity} on five tasks: \textbf{Looping\_Place-Retrieval}, \textbf{Sequential\_Picking}, \textbf{Bottle\_Handover}, \textbf{Food\_Replacement}, and \textbf{Cover\_and\_Reveal}, with details provided in Appendix Sec.~3.

\subsubsection{Implementation Details}
SeedPolicy uses a three-step observation history ($T_{\text{obs}}=3$), with $320\times240$ RGB images and 14-DoF joint poses. The latent state has length $N_s=60$ and dimension $D=256$. We train with AdamW using a batch size of 128, an initial learning rate of $10^{-4}$, cosine decay, and 500 warmup steps on a single NVIDIA RTX 4090D GPU. Owing to its parameter efficiency, the Transformer backbone is used by default for real-world experiments and ablations.

\begin{figure}[t]
    \centering
    \includegraphics[width= 0.95\linewidth]{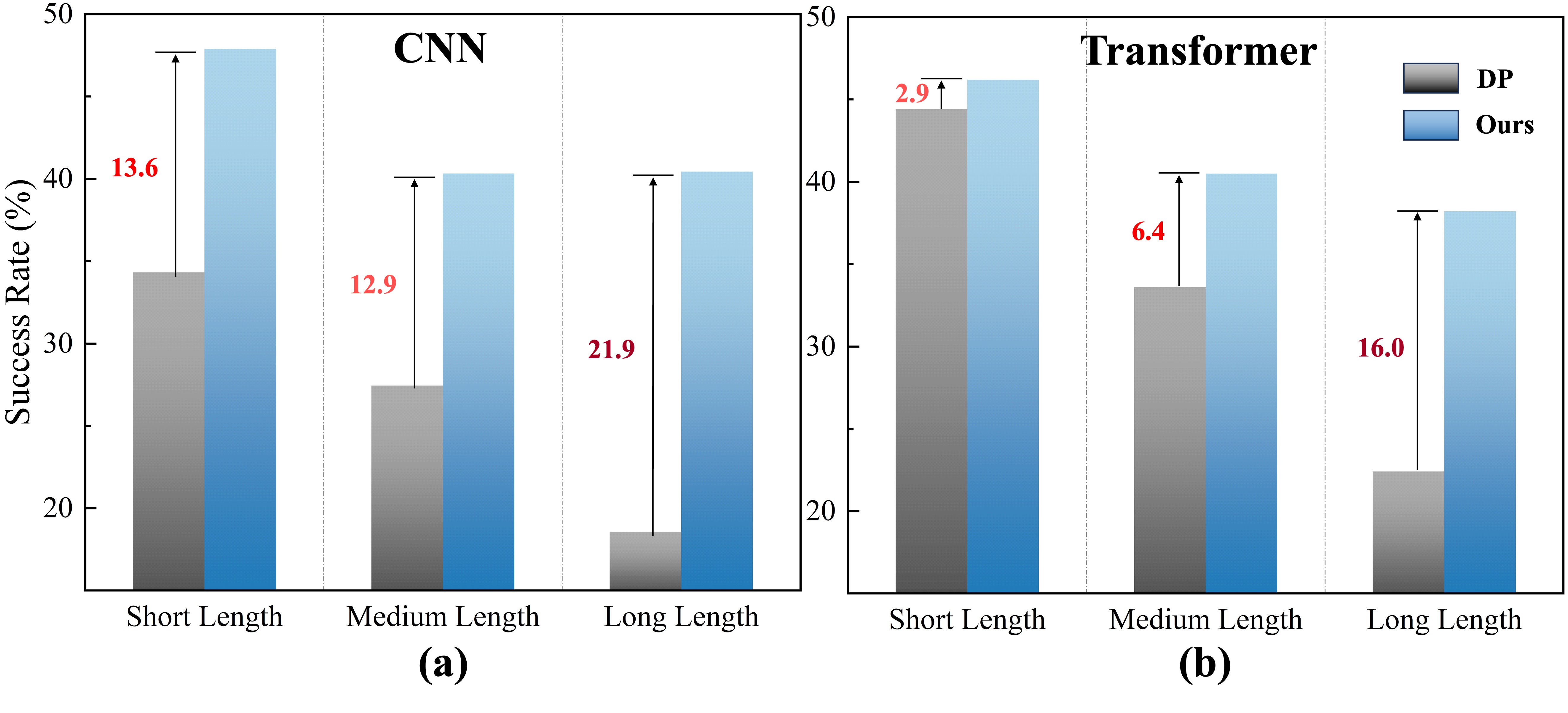}
    \caption{\textbf{Performance comparison across varying task length.} A consistent trend emerges in both architectures: as the task length increases, the performance gap between SeedPolicy and the baseline progressively widens. This validates the \textbf{architecture-agnostic effectiveness} of our approach, demonstrating that the advantage of our explicit temporal modeling becomes increasingly significant in long-horizon scenarios compared to fixed-window baselines.}
    \label{fig:difference}
    \vspace{-11pt}
\end{figure}

\begin{table*}[h]
\centering

\caption{Average success rates and parameter counts over 50 tasks. \textcolor{gray}{Gray indicates non-comparable VLA results.}}
\setlength{\tabcolsep}{5pt}
\footnotesize

\begin{tabular}{l >{\color{gray}}c | ccccc}
\toprule
\label{tab:1}

\multirow{2}{*}{\textbf{Method}} & \multirow{2}{*}{\textbf{RDT}~\cite{22}} & \multirow{2}{*}{\textbf{ACT}~\cite{1}} & \multicolumn{2}{c}{\textbf{DP~\cite{3}}} & \multicolumn{2}{c}{\textbf{SeedPolicy}} \\
\cmidrule(lr){4-5} \cmidrule(lr){6-7}
& & & Transformer & CNN & Transformer & CNN \\
\midrule

Avg. Success (Easy) & 34.50\% & 29.74\% & 33.10\% & 28.04\% & 40.08\% & \textbf{42.76\%} \\
Avg. Success (Hard) & 13.72\% & 1.74\% & 1.44\% & 0.64\% & \textbf{4.28\%} & 1.54\% \\
\midrule
\# Parameters & 1.2 B & 80 M & 20.61 M & 96.80 M & 33.36 M & 147.26 M \\
\bottomrule
\end{tabular}

\end{table*}

\begin{table*}[htbp] 
\centering
\caption{\textbf{Per-task performance comparison between Baseline and Ours on 50 tasks in  the "clean" setting.} Tasks are sorted by episode time (\textbf{Ep.Time}) (average duration of expert demonstrations in seconds). Higher success rates are highlighted in \textbf{bold}.}
\label{tab:2}
\renewcommand{\arraystretch}{1} 
\setlength{\tabcolsep}{2pt}       
\footnotesize
\resizebox{\linewidth}{!}{%
\begin{tabular}{l |c c |c c| c | l | c c| c c| c}
\toprule  
\multirow{2}{*}{\textbf{Task}} & \multicolumn{2}{c|}{\textbf{Transformer}} & \multicolumn{2}{c|}{\textbf{CNN}} & \multirow{2}{*}{\textbf{Ep.Time}} & 
\multirow{2}{*}{\textbf{Task}} & \multicolumn{2}{c|}{\textbf{Transformer}} & \multicolumn{2}{c|}{\textbf{CNN}} & \multirow{2}{*}{\textbf{Ep.Time}} \\

\cmidrule(lr){2-3} \cmidrule(lr){4-5} \cmidrule(lr){8-9} \cmidrule(lr){10-11}

 & \textbf{Baseline} & \textbf{Ours} & \textbf{Baseline} & \textbf{Ours} & & 
 & \textbf{Baseline} & \textbf{Ours} & \textbf{Baseline} & \textbf{Ours} & \\
\midrule 

Click Alarmclock          & 58 & \textbf{61} & 61 & \textbf{64} & 2  & Place Mouse Pad           & 0 & \textbf{1} & 0 & \textbf{1} & 5  \\
Click Bell                & 83 & \textbf{85} & 54 & \textbf{91} & 2  & Place Shoe                & 12 & \textbf{23} & 23 & \textbf{26} & 5  \\
Beat Block Hammer         & 72 & 72 & 42 & \textbf{61} & 3  & Rotate Qrcode              & 8 & \textbf{23} & 13 & \textbf{27} & 5  \\
Grab Roller               & 84 & \textbf{89} & \textbf{98} & 93 & 3  & Scan Object               & 2 & \textbf{9} & 9 & \textbf{11} & 5  \\
Lift Pot                  & \textbf{74} & 71 & 39 & \textbf{80} & 3  & Open Laptop               & \textbf{56} & 45 & \textbf{49} & 47 & 6  \\
Move Playingcard Away     & 49 & \textbf{68} & 47 & \textbf{58} & 3  & Handover Mic               & 83 & \textbf{92} & 53 & \textbf{95} & 7  \\
Turn Switch               & 51 & \textbf{54} & 36 & \textbf{49} & 3  & Place Bread Basket         & 4 & \textbf{20} & 14 & \textbf{21} & 7  \\
Adjust Bottle              & \textbf{97} & 83 & \textbf{97} & 92 & 4  & Place Dual Shoes           & 9 & \textbf{16} & 8 & 8 & 7  \\
Move Pillbottle Pad        & 2 & 2 & 1 & \textbf{6} & 4  & Place Burger Fries          & 63 & \textbf{66} & \textbf{72} & 43 & 8  \\
Pick Diverse Bottles        & \textbf{18} & 13 & 6 & \textbf{11} & 4  & Place Can Basket            & 34 & \textbf{40} & 18 & \textbf{65} & 8  \\
Pick Dual Bottles            & 24 & 24 & 24 & \textbf{29} & 4  & Place Object Basket          & 35 & \textbf{48} & 15 & \textbf{59} & 8  \\
Place Object Scale            & 0 & \textbf{5} & 1 & \textbf{7} & 4  & Shake Bottle                 & 90 & \textbf{93} & 65 & \textbf{94} & 8  \\
Place Object Stand            & 17 & \textbf{28} & 22 & \textbf{35} & 4  & Handover Block                & 14 & \textbf{21} & 10 & \textbf{51} & 9  \\
Place Phone Stand              & 4 & \textbf{13} & 13 & \textbf{23} & 4  & Place Cans Plasticbox         & \textbf{47} & 18 & \textbf{40} & 29 & 9  \\
Press Stapler                   & 70 & \textbf{76} & 6 & \textbf{56} & 4  & Put Object Cabinet             & 11 & \textbf{41} & 42 & \textbf{43} & 9  \\
Stamp Seal                      & 3 & \textbf{8} & 2 & \textbf{11} & 4  & Shake Bottle Horizontally      & 95 & 95 & 59 & \textbf{96} & 9  \\
Dump Bin Bigbin                 & 41 & \textbf{52} & 49 & \textbf{55} & 5  & Stack Blocks Two                & 28 & \textbf{47} & 7 & \textbf{58} & 10 \\
Move Can Pot                    & 69 & \textbf{71} & 39 & \textbf{75} & 5  & Stack Bowls Two                 & 33 & \textbf{73} & 61 & \textbf{76} & 10 \\
Move Stapler Pad                & 0 & 0 & 1 & 1 & 5  & Hanging Mug                     & 7 & \textbf{9} & 8 & 8 & 11 \\
Place A2B Left                  & 4 & \textbf{9} & 2 & \textbf{18} & 5  & Open Microwave                  & 79 & \textbf{80} & 5 & \textbf{84} & 14 \\
Place A2B Right                 & 2 & \textbf{12} & \textbf{13} & 11 & 5  & Blocks Ranking RGB              & 4 & 4 & 0 & \textbf{8} & 15 \\
Place Bread Skillet             & 9 & \textbf{14} & 11 & \textbf{16} & 5  & Blocks Ranking Size             & 0 & \textbf{3} & 1 & \textbf{3} & 15 \\
Place Container Plate            & 29 & \textbf{60} & 41 & \textbf{52} & 5  & Stack Blocks Three               & 10 & \textbf{15} & 0 & \textbf{17} & 15 \\
Place Empty Cup                  & 24 & \textbf{32} & 37 & \textbf{50} & 5  & Stack Bowls Three                & 25 & \textbf{67} & 63 & \textbf{72} & 15 \\
Place Fan                        & 5 & 5 & 3 & \textbf{14} & 5  & Put Bottles Dustbin               & 16 & \textbf{48} & 22 & \textbf{38} & 20 \\

\bottomrule
\end{tabular}
}
\vspace{-8pt}
\end{table*}

\subsection{Results}

\textbf{SeedPolicy delivers consistent performance gains and robustness across tasks (Q1).} On the RoboTwin 2.0, SeedPolicy outperforms or matches the DP baseline on \textbf{45/50 tasks} with the Transformer backbone and \textbf{44/50 tasks} with the CNN backbone, demonstrating architecture-agnostic effectiveness.

In the ``Easy'' setting, SeedPolicy improves DP by \textbf{7.0\%} absolute / \textbf{21.10\%} relative with the Transformer backbone, and by \textbf{14.72\%} absolute / \textbf{52.50\%} relative with the CNN backbone, indicating that SEGA effectively compensates for the limited temporal modeling capability of DP.

\begin{figure*}[t]
    \centering
    \includegraphics[width= 0.9\linewidth]{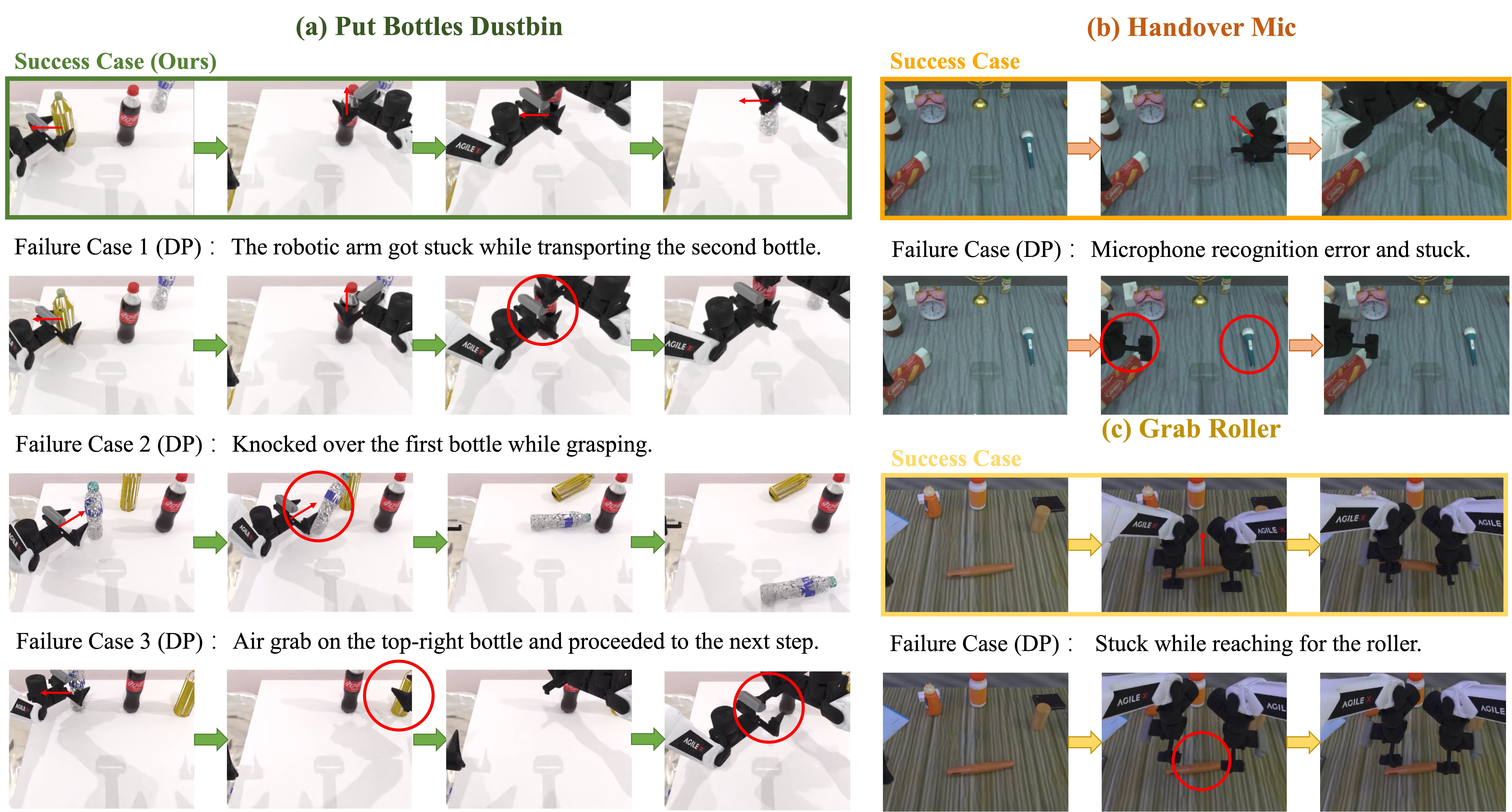}
    \caption{\textbf{Qualitative failure cases in simulation.} We compare the successful execution of SeedPolicy (top row) against representative failure modes of the DP across three tasks: (a) \textit{Put Bottles Dustbin} ("clean" setting), (b) \textit{Handover Mic} ("hard" setting), and (c) \textit{Grab Roller} ("hard" setting). Red circles highlight critical errors, including execution stagnation (getting stuck) and spatial positioning failures (collisions or air grabs). Additional visualizations are provided in the Appendix.}
    \label{fig:simulation failcase}
    \vspace{-8pt}
\end{figure*}

In the Hard setting, SeedPolicy improves over DP from 1.44\% to 4.28\% with the Transformer backbone and from 0.64\% to 1.54\% with the CNN backbone, showing consistent gains under severe randomization despite the challenging nature of this setting.

Efficiency-wise, SeedPolicy achieves stronger clean-setting performance than the reported foundation model-based RDT~\cite{22} result, while using substantially fewer parameters (SeedPolicy-Transformer: 33.36M, SeedPolicy-CNN: 147.26M vs. RDT's 1.2B). In the ``Easy'' setting, SeedPolicy's CNN backbone achieves a \textbf{42.76\% success rate}, surpassing RDT by \textbf{over 8 percentage points}. Although RDT performs better in the ``Hard'' setting (\textbf{13.72\%}), this is expected since vision-language-action models benefit from large-scale pre-trained vision and language encoders for open-ended generalization.

\begin{table}[t]
    \centering
    \caption{Real-world success rates (\%) across five tasks.}
    \label{tab:real_world_results}
    \small
    \setlength{\tabcolsep}{4pt}
    \renewcommand{\arraystretch}{1}
    \begin{tabularx}{\columnwidth}{
        >{\raggedright\arraybackslash}X
        >{\centering\arraybackslash}p{0.15\columnwidth}
        >{\centering\arraybackslash}p{0.19\columnwidth}
    }
        \toprule
        \textbf{Task} & \textbf{DP} & \textbf{SeedPolicy} \\
        \midrule
        Looping Place-Retrieval & 14 & \textbf{38} \\
        Sequential Picking      & 10 & \textbf{26} \\
        Bottle Handover         & 16 & \textbf{54} \\
        Food Replacement        & 18 & \textbf{60} \\
        Cover and Reveal        & 16 & \textbf{58} \\
        \bottomrule
    \end{tabularx}
    \vspace{-12pt}
\end{table}

\textbf{SeedPolicy exhibits stronger long-term dependency modeling, with larger gains on longer tasks (Q2).} 
To evaluate temporal reasoning, we divide the 50 tasks into Short, Medium, and Long groups according to their average episode steps, and compare success rates for both CNN and Transformer backbones in Fig.~\ref{fig:difference}. 

A consistent trend emerges across architectures: as task length increases, the performance margin of SeedPolicy over DP becomes larger.
For short tasks, the fixed-window baseline is still relatively sufficient, leading to a modest \textbf{2.9\%} gain for the Transformer backbone, while the CNN backbone benefits more substantially from explicit temporal modeling with a \textbf{13.6\%} gain. 
For medium-length tasks, the gap further increases to \textbf{6.4\%} for Transformer and remains large at \textbf{12.9\%} for CNN, indicating that SeedPolicy better preserves intermediate task states as temporal complexity grows. 
The largest improvement appears in long-horizon tasks, where the baseline suffers from limited-window information loss, while SeedPolicy maintains robust performance and widens the margin to \textbf{16.0\%} for Transformer and \textbf{21.9\%} for CNN.

These results support our hypothesis that SEGA better tracks progress in multi-stage tasks by maintaining an evolving latent state, whereas fixed-window policies struggle to retain historical context in extended tasks.

\begin{figure*}[t]
    \centering
    \includegraphics[width= 0.9\linewidth]{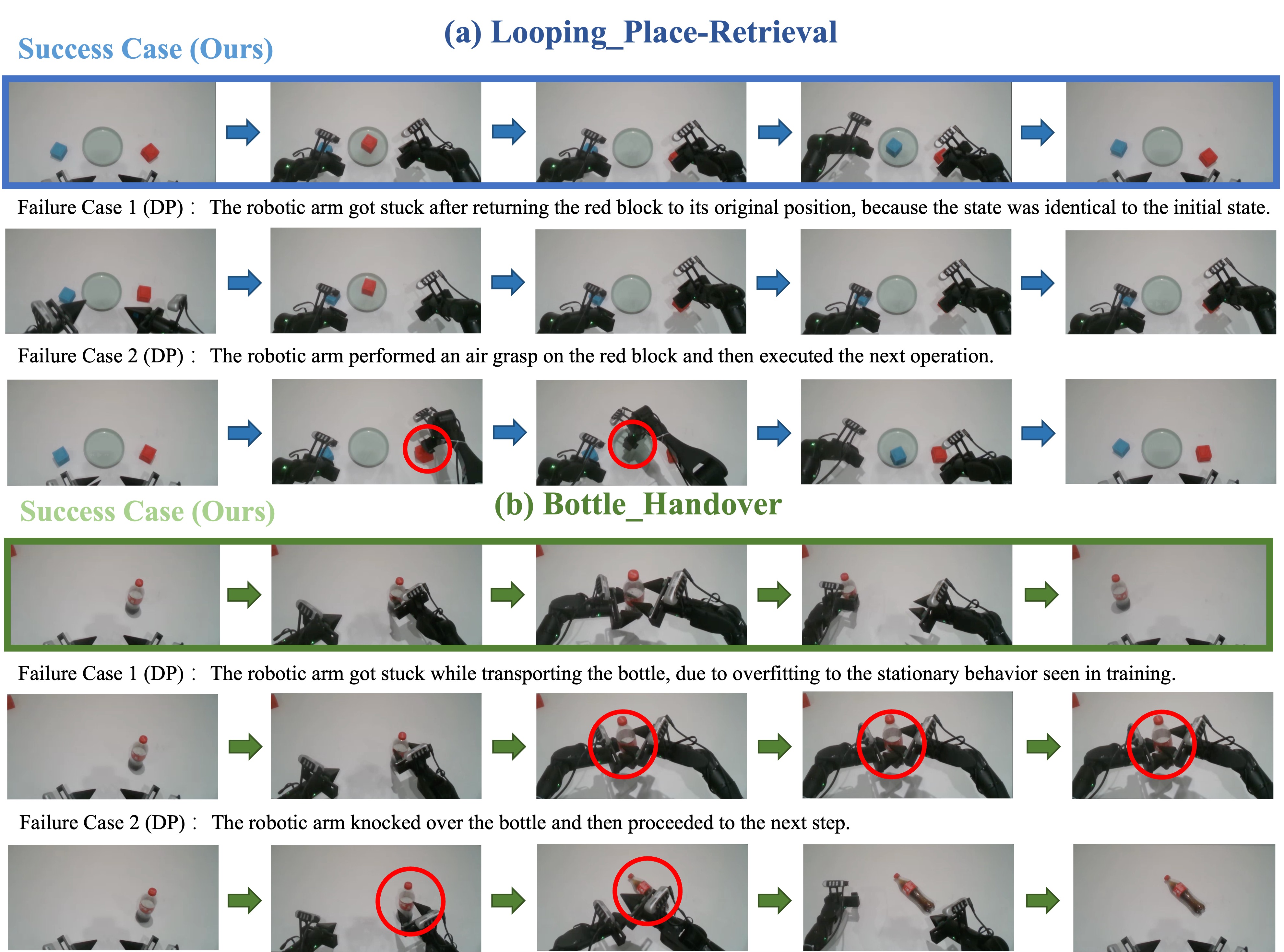}
    \caption{\textbf{Qualitative failure analysis in real-world scenarios.} We visualize the successful execution of SeedPolicy (top rows) compared to common baseline failures in (a) \textit{Looping\_Place-Retrieval} and (b) \textit{Bottle\_Handover}. Red circles highlight critical errors. 
    \textbf{Failure Case 1} illustrates execution stagnation caused by \textit{Perceptual Aliasing} (misinterpreting the returned block as the initial state in (a)) or overfitting to pauses (in (b)). \textbf{Failure Case 2} demonstrates spatial precision errors (air grabs and collisions) attributed to the lack of explicit depth information. Additional visualization analysis is provided in the Appendix.}
    \label{fig:real world failcase}
    \vspace{-8pt}
\end{figure*}

\textbf{SeedPolicy is more robust to execution stagnation and spatial positioning errors (Q2).}
We visualize representative failure cases in simulation (Fig.~\ref{fig:simulation failcase}) and the real-world (Fig.~\ref{fig:real world failcase}), which explain the quantitative gains in Table~\ref{tab:real_world_results}, e.g., improving \textit{Bottle\_Handover} from 16\% to 54\%. 
The baseline mainly suffers from two failure modes.

\noindent \textbf{1) Execution stagnation and state aliasing.}
In multi-stage tasks such as \textit{Put Bottles Dustbin}, \textit{Looping\_Place-Retrieval}, and \textit{Bottle\_Handover}, demonstrations often contain pauses or visually repeated states. 
This creates phase ambiguity for fixed-window policies: the baseline may freeze after returning to a state visually similar to the start, or overfit to stationary pauses and fall into a zero-velocity loop. 
SeedPolicy mitigates this issue by recursively updating the latent state, which preserves task progress and enables temporally consistent transitions across ambiguous phases.

\begin{table}[t]
\centering
\caption{Comparison of different memory mechanisms on ten manipulation tasks.}
\label{tab:memory_compare}
\scriptsize
\setlength{\tabcolsep}{2pt}
\renewcommand{\arraystretch}{0.85}
\resizebox{\linewidth}{!}{
\begin{tabular}{lccc}
\toprule
Task & DP + ARMT-style & DP + MemoryVLA-style & SeedPolicy \\
\midrule
Move Playingcard Away         & 56 & 64 & \textbf{68} \\
Turn Switch                   & 50 & 52 & \textbf{54} \\
Place Object Stand            & 20 & 26 & \textbf{28} \\
Dump Bin Bigbin               & 47 & 50 & \textbf{52} \\
Place Container Plate         & 38 & 51 & \textbf{60} \\
Place Empty Cup               & 15 & 30 & \textbf{32} \\
Put Object Cabinet            & 15 & 29 & \textbf{41} \\
Stack Blocks Two              & 33 & 38 & \textbf{47} \\
Stack Bowls Two               & 56 & 60 & \textbf{73} \\
Put Bottles Dustbin           & 21 & 26 & \textbf{48} \\
\bottomrule
\end{tabular}
}
\vspace{-10pt}
\end{table}

\noindent \textbf{2) Spatial positioning errors without depth.} 
Across both simulation and real-world cases, the baseline frequently exhibits air grabs or collisions, mainly because precise 3D object localization from fixed-view RGB observations is ambiguous. 
With only a limited observation window, DP struggles to resolve this depth uncertainty. 
In contrast, SeedPolicy accumulates long-horizon motion cues in its evolving latent state, enabling better spatial inference from historical trajectories and fewer positioning errors.

\textbf{SEGA outperforms existing memory mechanisms when integrated into Diffusion Policy (Q3).}

We construct two controlled DP variants by replacing SEGA with ARMT-style~\cite{47} recurrent memory and MemoryVLA-style~\cite{46} external memory modules, and evaluate all variants on ten tasks under the same DP backbone and training protocol. Specifically, ARMT-style~\cite{47} recurrent transformers organize memory retrieval and update in a single recurrent chain. While MemoryVLA-style~\cite{46} also performs retrieval, its update mainly integrates fused representations into an external memory bank. In contrast, SEGA explicitly performs latent-state evolution and observation enhancement through two parallel state--observation interaction streams at each timestep. As shown in Table~\ref{tab:memory_compare}, SeedPolicy consistently achieves the best performance, especially on the long-horizon task \textit{Stack Bowls Two} and \textit{Put Bottles Dustbin}. These results suggest that the explicit dual-stream interaction in SEGA is more effective for long-horizon manipulation than directly adopting existing memory mechanisms in Diffusion Policy.

\begin{table}[t]

\centering
\renewcommand{\arraystretch}{1} 
\setlength{\tabcolsep}{4pt}       
\caption{Component-wise Ablation of SeedPolicy.}
\label{tab:ablation}

\resizebox{\linewidth}{!}{ 
\begin{tabular}{@{}l c c c c c@{}}
\toprule
\multirow{2}{*}{\textbf{Task}} & \multicolumn{4}{c}{\textbf{Architecture Design}} \\
\cmidrule(lr){2-6}
& DP & + Temp. Att. & + State & \multicolumn{2}{c}{+ Gating} \\
\cmidrule(lr){5-6}
& & & & CA (SeedPolicy) & FFN \\
\midrule
Turn Switch (Short)         & 51 & 51 & 51 & \textbf{54} & 53 \\
Place Empty Cup (Medium)    & 24 & 26 & 28 & \textbf{32} & 21 \\
Stack Bowls Two (Long)      & 33 & 48 & 65 & \textbf{73} & 70 \\
\bottomrule
\end{tabular}
}
\vspace{-10pt}
\end{table}

\textbf{Both recurrent state modeling and cross-attention-based gating are critical for effective horizon scaling (Q4).}
Table~\ref{tab:ablation} progressively ablates DP, DP with temporal attention, DP with the recurrent State mechanism (no gate), and different gating designs. 
For the short-horizon task \textit{Turn Switch}, all temporal variants perform similarly ($\sim51\%$), while SeedPolicy slightly improves to $54\%$, indicating that gating also benefits precise short-term control. 
As the horizon increases, the temporal bottleneck of frame stacking becomes clearer: on \textit{Stack Bowls Two}, DP achieves only $33\%$, while temporal attention improves it to $48\%$. 
Replacing temporal attention with our recurrent State mechanism further raises the success rate to $65\%$, showing that a compact evolving state is more effective for long-horizon context modeling while avoiding the quadratic cost of long-window attention.

Adding SEG achieves the best performance, reaching $54\%$, $32\%$, and $73\%$ on short-, medium-, and long-length tasks, respectively. 
Compared with a generic FFN gate, the cross-attention-based gate performs more consistently, especially on the medium task \textit{Place Empty Cup} ($32\%$ vs. $21\%$) and the long task \textit{Stack Bowls Two} ($73\%$ vs. $70\%$). This suggests that intrinsic cross-attention logits provide more reliable relevance signals than purely learnable gating. 

\begin{figure}[t]
    \centering
    
    \captionsetup{font=scriptsize}
    
    \includegraphics[width=\linewidth]{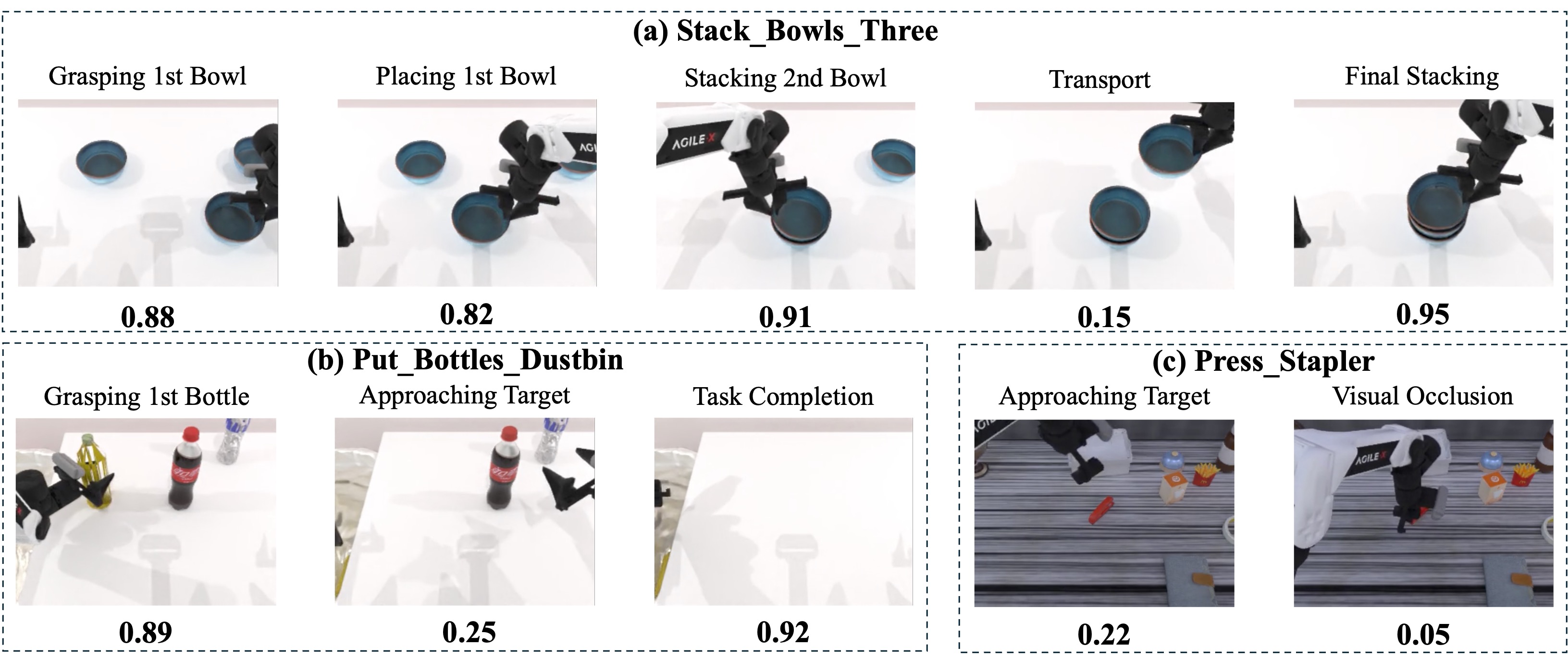}
    \setlength{\abovecaptionskip}{2pt}
    \setlength{\belowcaptionskip}{0pt}
    \caption{\textbf{Visualization Analysis of the Self-Evolving Gating Mechanism.}}
    \label{fig:gate_analysis}
    \vspace{-10pt}
\end{figure}

Fig.~\ref{fig:gate_analysis} further supports this interpretation through visualizations of average gate values during execution, with the numbers below each image indicating the corresponding values. In (a) and (b), the gates peak at semantic interaction points, such as grasping and stacking, and decrease during transport and approach, reflecting temporal sparsity. 
In (c), the gate maintains state stability by filtering out irrelevant frames during occlusions.

\section{Limitations and Future Work}
\label{sec:limitations and future work}

Appendix~\ref{app:experiment} provides preliminary evidence that the recurrent latent-state mechanism is compatible with VLA-style architectures, suggesting its potential beyond standard diffusion policies. 
Moreover, current World Action Models (WAMs) still face challenges in robust long-horizon execution, motivating future integration of SEGA-style recurrent states for stronger temporal context modeling.

\section{Conclusion} 
\label{sec:conclusion}

SeedPolicy addresses the horizon scaling bottleneck in imitation learning through Self-Evolving Gated Attention (SEGA), which maintains a compact, recursively updated latent state to extend the effective temporal receptive field without increasing the per-step observation window. By further using cross-attention-based gating to filter noisy observations and preserve task-relevant context, SEGA enables efficient and robust temporal modeling for long-horizon manipulation. Extensive evaluations demonstrate that SeedPolicy consistently improves over Diffusion Policy and remains competitive with billion-parameter vision-language-action models while using significantly fewer parameters. These results offer a promising direction for efficient long-horizon robotic manipulation.

\clearpage

\bibliography{references}

\clearpage
\appendix

\section{Appendix} 
\label{sec:appendix}

\subsection{Overview}

In this appendix, we provide an extended discussion of related work, supplementary details and extensive experimental results to further substantiate the effectiveness and robustness of the proposed SeedPolicy. The organization of the appendix is summarized in Table~\ref{tab:appendix_overview}.

\begin{table}[h]
    \centering
    \caption{\textbf{Overview of the Appendix Content.} This table outlines the supplementary materials provided to support the main paper.}
    \label{tab:appendix_overview}
    \renewcommand{\arraystretch}{1.4} 
    \resizebox{\linewidth}{!}{%
    \begin{tabular}{p{0.3\linewidth} p{0.65\linewidth}}
        \toprule
        \textbf{Section} & \textbf{Description} \\
        \midrule

        \textbf{Related Work} & An extended review of prior studies related to imitation learning for robotic manipulation and gating mechanisms in sequence modeling. \\

        \textbf{Hyperparameter Analysis} & 
        A comprehensive sensitivity analysis on the key hyperparameters of the Self-Evolving Gated Attention (SEGA) module. \\
        
        \textbf{Hardware \& Data Collection and Processing} & 
        Detailed specifications of the hardware configurations (DOS-W1) and the data collection and processing pipeline used to curate expert demonstrations. \\
        
        \textbf{Additional Results} & 
        Additional quantitative and qualitative comparisons on both the RoboTwin 2.0 simulation benchmark and real-world robotic setups. \\
        
        \textbf{Open-Loop Evaluation} & 
        Open-loop evaluations on the training set to verify the model's capacity and trajectory reconstruction quality. \\
        \bottomrule
    \end{tabular}
    }
\end{table}

\section{Related Work}
\label{sec:Realted Work}

\subsection{Imitation Learning for Robotic Manipulation}
Imitation Learning (IL) learns observation-action mappings from expert demonstrations and has become a dominant paradigm for robotic manipulation. 
Early Behavior Cloning (BC) methods directly map observations to actions but often suffer from covariate shift and compounding errors. 
Recent works have explored sequence modeling~\cite{28, 29, 35,36,37, 39} and generative policies~\cite{30, 31, 32,33,34,38,40,41} to improve temporal consistency and capture multi-modal behaviors.

Representative sequence-based methods include \textbf{ACT}~\cite{1}, which predicts action chunks using a CVAE-based architecture, and \textbf{BeT}~\cite{2}, which discretizes actions and models multi-modal behavior with a GPT-style architecture. 
More recently, generative policies have achieved strong performance by explicitly modeling the distribution of expert actions. 
\textbf{Diffusion Policy}~\cite{3} formulates visuomotor control as a conditional denoising process over action sequences, while \textbf{DP3}~\cite{4} extends diffusion policies to 3D point cloud inputs for better spatial generalization. 
Flow-based models such as \textbf{G3Flow}~\cite{5} further improve inference efficiency through flow matching.

However, most existing methods only rely on fixed-length observation histories, such as frame stacking, which limits their ability to capture long-term temporal dependencies. 
In contrast, our framework employs a recursively updating latent state to effectively capture long-term spatiotemporal dependencies, allowing the policy to scale its effective temporal receptive field without relying on longer frame stacks.

\subsection{Gating Mechanisms in Sequence Modeling}

Gating mechanisms are widely used to regulate information flow in neural networks. 
Early recurrent architectures, such as LSTMs~\cite{6} and GRUs~\cite{7}, introduced learnable gates to selectively retain or discard temporal information. 
This idea was later extended to feedforward depth via Highway Networks~\cite{8} and to Transformer architectures, where GLU variants such as SwiGLU~\cite{9} have become standard components for improving expressiveness and training stability~\cite{10, 11,12, 43, 45}.

Recent studies further apply data-dependent gates to attention outputs. 
For example, \citet{19} proposed Gated Attention, showing that gating on SDPA outputs introduces non-linearity, mitigates attention sinks~\cite{42}, and enables more precise context filtering. 
This capability is particularly important for robotic manipulation, where visual streams often contain redundant or noisy observations, such as background shifts, and occlusions. 
Since standard softmax attention can be sensitive to such noise~\cite{zhou2024multimax}, our Self-Evolving Gate (SEG) leverages cross-attention logits as relevance signals to enforce semantic sparsity, preserving informative historical features while suppressing distractions.

\subsection{Hyperparameter Analysis}
\paragraph{Impact of Latent State Sequence Length ($N_s$)}

We investigate the impact of the latent state sequence length $N_s$, which governs the temporal capacity of the SEGA module. 

As detailed in Table~\ref{tab:ablation_ns}, setting $N_s=30$ results in suboptimal performance, particularly in multi-stage tasks like \textit{Put Object Cabinet} ($32\%$) and \textit{Stack Bowls Two} ($56\%$). This suggests that a shorter history fails to retain sufficient context for resolving long-horizon dependencies and state ambiguities.

Increasing the length to $N_s=60$ yields the best performance across all representative tasks, indicating an optimal balance where the model effectively captures necessary historical cues.

However, further extending the length to $N_s=90$ leads to a performance plateau or slight degradation (e.g., \textit{Grab Roller} drops from $89\%$ to $80\%$). This decline implies that an excessively long history may distract the attention mechanism from critical signals. 
Consequently, we adopt $N_s=60$ as the default setting.

\begin{table}[htbp]
\centering
\caption{\textbf{Ablation study on latent state sequence length ($N_s$).} We report the success rates (\%) on six representative tasks from RoboTwin 2.0. The results demonstrate that $N_s=60$ achieves superior performance while maintaining a smaller parameter size compared to $N_s=90$.}
\label{tab:ablation_ns}
\renewcommand{\arraystretch}{1.2} 
\setlength{\tabcolsep}{10pt}      

\begin{tabular}{l c c c}
\toprule
\multirow{2}{*}{\textbf{Task}} & \multicolumn{3}{c}{\textbf{State}} \\
\cmidrule(lr){2-4}
& 30 & \textbf{60} & 90 \\
\midrule
Grab Roller         & 74  & \textbf{89}  & 80  \\
Dump Bin Bigbin     & 43 & \textbf{52} & 47 \\
Open Microwave      & 73 & \textbf{80} & 79 \\
Handover Mic        & 80 & \textbf{92} & 86 \\
Stack Bowls Two     & 56 & \textbf{73} & \textbf{73} \\
Put Object Cabinet  & 32 & \textbf{41} & 36 \\
\bottomrule
\end{tabular}
\end{table}

\paragraph{Depth of SEGA Interaction Blocks ($L$)} 

We examine the effect of the network depth (number of attention blocks) on policy performance.

As presented in Table~\ref{tab:ablation_depth}, increasing the depth from 2 to 6 yields consistent performance gains across all evaluated tasks. Notably, complex tasks such as \textit{Put Bottles Dustbin} see a significant boost (from 36\% at depth 2 to 48\% at depth 6), underscoring the need for sufficient model capacity to encode intricate manipulation behaviors.

However, further increasing the depth to 8 leads to a marked performance regression (e.g., \textit{Move Can Pot} drops from 71\% to 52\%). This degradation is likely attributed to overfitting, as the larger parameter space becomes harder to regularize given the limited number of expert demonstrations (50 per task).

Thus, a depth of 6 provides the optimal balance between expressivity and generalization.

\begin{table}[htbp]
\centering

\caption{Ablation study on the depth of attention blocks. We report the success rate (\%) on seven representative tasks. The model achieves the best balance and performance at depth 6.}
\label{tab:ablation_depth}
\renewcommand{\arraystretch}{1.2} 
\setlength{\tabcolsep}{10pt}      

\begin{tabular}{l c c c c}
\toprule
\multirow{2}{*}{\textbf{Task}} & \multicolumn{4}{c}{\textbf{Network Depth}} \\
\cmidrule(lr){2-5}
 & 2 & 4 & \textbf{6} & 8 \\
\midrule
Scan Object         & 3  & 2  & \textbf{9}  & 2  \\
Beat Block Hammer   & 56 & 35 & \textbf{72} & 47 \\
Move Can Pot        & 59 & 63 & \textbf{71} & 52 \\
Dump Bin Bigbin     & 44 & 43 & \textbf{52} & 44 \\
Handover Mic        & 83 & 77 & \textbf{92} & 86 \\
Put Object Cabinet  & 34 & 37 & \textbf{41} & 30 \\
Put Bottles Dustbin & 36  & 47  & \textbf{48}  & 33  \\
\bottomrule
\end{tabular}
\end{table}

\paragraph{Training Hyperparameter Details}
Table~\ref{tab:hyperparameters} summarizes the hyperparameter settings for SeedPolicy. We train all models using the AdamW optimizer for 600 epochs with a batch size of 128. A learning rate of $1 \times 10^{-4}$ is used with a cosine decay scheduler and a 500-step warmup.

Distinct optimization parameters are applied to match the inductive biases of the backbones: the \textbf{Transformer variant} uses a higher weight decay ($1e^{-3}$) for its attention layers and adjusted $\beta$ parameters $(0.9, 0.95)$, whereas the \textbf{CNN variant} uses a standard configuration.

For the diffusion backend, we employ a DDPM scheduler with 100 training and inference timesteps, utilizing a Squared Cosine Cap v2 beta schedule ($\beta_{\text{start}}=1e^{-4}, \beta_{\text{end}}=0.02$) and $\epsilon$-prediction.

\begin{table}[h]
    \centering
    \caption{\textbf{Hyperparameter configurations for SeedPolicy.} We report the settings for both Transformer and CNN backbones used in our experiments.}
    \label{tab:hyperparameters}
    \resizebox{0.95\linewidth}{!}{%
    \begin{tabular}{l|cc}
        \toprule
        \textbf{Hyperparameter} & \textbf{SeedPolicy-Transformer} & \textbf{SeedPolicy-CNN} \\
        \midrule
        \multicolumn{3}{c}{\textit{Optimization \& Training}} \\
        \midrule
        Optimizer & AdamW & AdamW \\
        Learning Rate & $1 \times 10^{-4}$ & $1 \times 10^{-4}$ \\
        LR Scheduler & Cosine Decay & Cosine Decay \\
        LR Warmup Steps & 500 & 500 \\
        Batch Size & 128 & 128 \\
        Weight Decay & $1e^{-3}$ (Trans.), $1e^{-6}$ (Enc.) & $1 \times 10^{-6}$ \\
        Betas ($\beta_1, \beta_2$) & $(0.9, 0.95)$ & $(0.95, 0.999)$ \\
        Num Epochs & 600 & 600 \\
        EMA Decay & 0.75 & 0.75 \\
        Gradient Accumulation & 1 & 1 \\
        \midrule
        \multicolumn{3}{c}{\textit{Diffusion \& Action Process}} \\
        \midrule
        Noise Scheduler & DDPM & DDPM \\
        Training Timesteps & 100 & 100 \\
        Inference Timesteps & 100 & 100 \\
        Beta Schedule & Squared Cosine Cap v2 & Squared Cosine Cap v2 \\
        Beta Range & $1e^{-4} \to 0.02$ & $1e^{-4} \to 0.02$ \\
        Prediction Type & Epsilon ($\epsilon$) & Epsilon ($\epsilon$) \\
        \bottomrule
    \end{tabular}
    }
\end{table}

\subsection{Hardware Setup and Data Collection and Processing}
\label{app:data}

\begin{figure}[t]
    \vspace{-10pt}
    \centering
    \includegraphics[width= 0.5\linewidth]{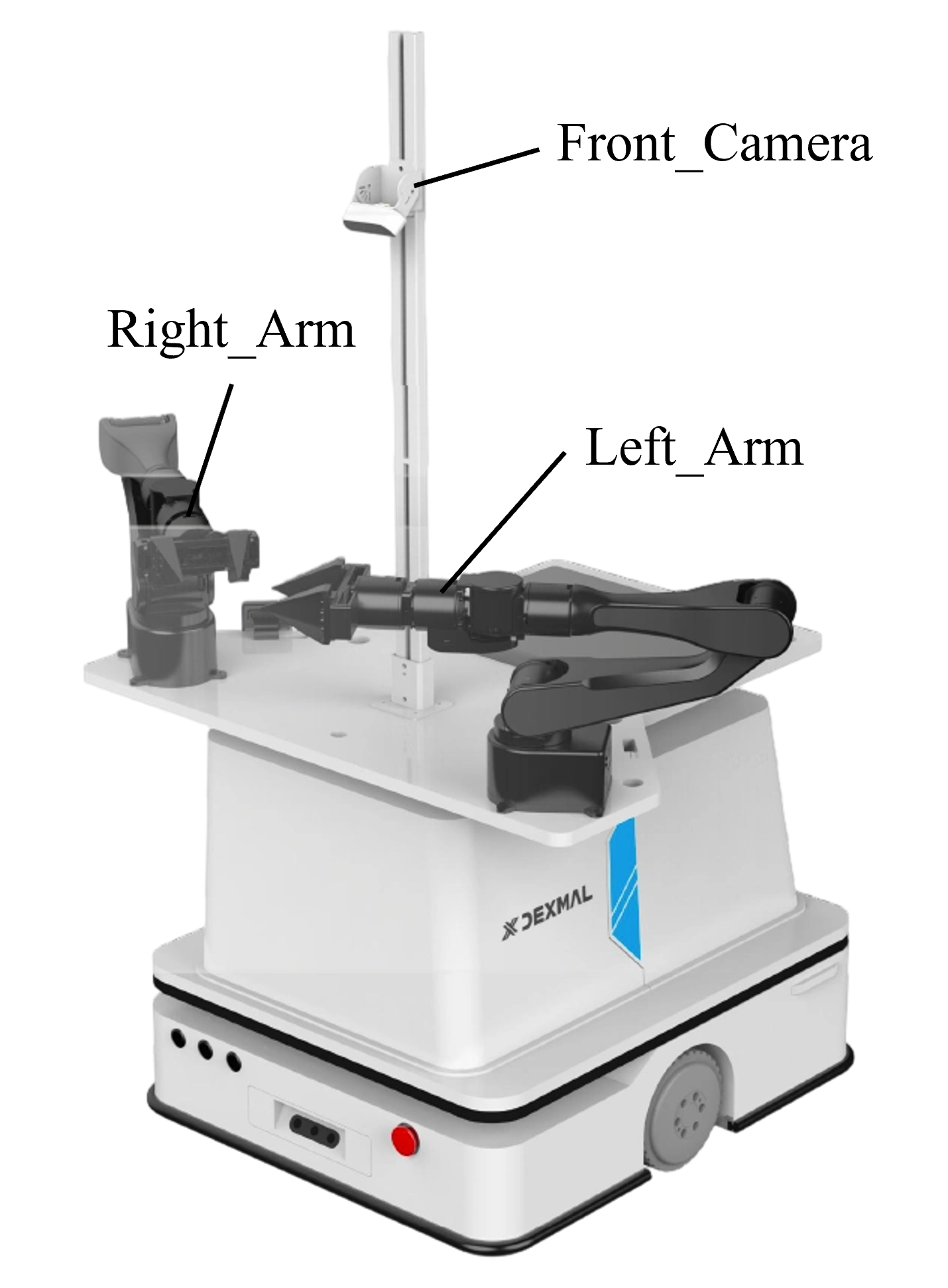}
    \caption{\textbf{The DOS-W1 Mobile Manipulation Platform.} As illustrated, the system integrates dual 7-DoF robotic arms with a differential drive mobile chassis and a vertical lift mechanism. A front-view RGB camera is mounted on the mast for visual perception.}
    \label{fig:Dos_W1}
    \vspace{-10pt}
\end{figure}

\noindent \textbf{Robotic Platform Specifications.} We utilize the \textbf{DOS-W1} (as shown in Fig.~\ref{fig:Dos_W1}), a dual-arm mobile manipulation platform developed by Dexmal. The system features a highly articulated design with a total of 17 degrees of freedom (DoF), comprising two 7-DoF robotic arms (6-DoF arm + 1-DoF gripper) and a mobile chassis equipped with a differential drive and a vertical lift mechanism. The dual arms offer a payload capacity of 1.5 kg each with a repeatability of $\pm 0.1$ mm, ensuring precise manipulation capabilities. The mobile base supports a heavy payload of 300 kg and includes an adjustable lift range of 600--880 mm to adapt to varying workspace heights. Detailed hardware specifications are summarized in Table~\ref{tab:hardware_specs}.

\begin{table}[h]
    \centering
    \caption{\textbf{Hardware Specifications of the DOS-W1 Platform.} The system integrates dual 7-DoF manipulators with a high-payload differential mobile base, featuring an adjustable lift mechanism for versatile workspace adaptation.}
    \label{tab:hardware_specs}
    \resizebox{0.95\linewidth}{!}{%
    \begin{tabular}{l l c}
        \toprule
        \textbf{Component} & \textbf{Parameter} & \textbf{Value} \\
        \midrule
        \multirow{5}{*}{\textbf{Dual Arms}} 
         & Degrees of Freedom (DoF) & $2 \times (6 \text{ Arm} + 1 \text{ Gripper})$ \\
         & Payload Capacity & 1.5 kg / arm \\
         & Working Radius & 647 mm \\
         & Repeatability & $\pm 0.1$ mm \\
        \midrule
        \multirow{5}{*}{\textbf{Mobile Chassis}} 
         & Drive Type & Differential Drive \\
         & Dimensions ($L \times W$) & $700 \times 620$ mm \\
         & Vertical Lift Stroke & $600 \sim 880$ mm (Desktop Height) \\
         & Max. Payload & 300 kg \\
         & Chassis Self-Weight & 150 kg \\
        \bottomrule
    \end{tabular}
    }
\end{table}

\noindent \textbf{Real-World Data Collection and Processing.} To ensure policy robustness and bridge the gap between raw logs and training formats, we implemented a comprehensive pipeline covering diversity-aware collection, temporal alignment, and efficient storage.

\vspace{0.5em}

\noindent \textit{1) Task Setup and Randomization:} We collected 50 expert demonstrations for five representative long-horizon tasks: \textit{Sequential\_Picking}, \textit{Bottle\_Handover}, \textit{Looping\_Place-Retrieval}, \textit{Food\_Replacement}, and \textit{Cover\_and\_Reveal}. To foster strong generalization capabilities, we introduced systematic spatial randomization during the data collection phase. For each episode, we explicitly varied the initial positions and placement poses (orientations) of the target objects. This randomization strategy ensures that the policy learns to perceive relative spatial geometry rather than overfitting to fixed absolute coordinates.

\vspace{0.5em}
\noindent \textit{2) Temporal Alignment and Static Filtering:} Following collection, we address the asynchronous nature of the raw data. The raw data from the DOS-W1 platform comprises independent streams of RGB video and high-frequency proprioceptive logs. We first synchronize these streams by aligning proprioceptive timestamps to video frame timestamps via nearest-neighbor matching. To ensure data quality, we employ a motion-based filtering mechanism: a frame at time $t$ is discarded if the 14-dimensional state vector (comprising 6-DoF joint angles and 1-DoF gripper states for both arms) remains static compared to the previous valid frame, defined by an absolute tolerance of $\epsilon < 1 \times 10^{-4}$.

\vspace{0.5em}
\noindent \textit{3) Dataset Aggregation and Formatting:} In the final stage, the processed episodes are aggregated into Zarr archives to optimize I/O throughput. We format the imitation learning objective as a next-state prediction problem, where the action $A_t$ corresponds to the state vector at $t+1$. Furthermore, visual observations are decoded and transposed to the channel-first (NCHW) format standard for deep learning, and the final dataset is compressed using the Blosc-Zstd algorithm (level 3) to balance storage efficiency with access speed.

\noindent \textbf{Description of Real-World Tasks.} To further evaluate the ability of SeedPolicy to handle state ambiguity and long-term task progression, we design five representative tasks with repeated visual states, ordered subgoals, or phase-dependent behaviors:

\begin{itemize}
    \item \textbf{Looping\_Place-Retrieval}: The robot places a red block into a tray and then retrieves it, followed by the same place-and-retrieve sequence for a blue block. This task introduces repeated visual configurations, requiring the policy to distinguish different execution phases.

    \item \textbf{Sequential\_Picking}: The robot picks and places red, yellow, and blue blocks in a predefined order. The task requires the policy to track task progress and avoid confusing visually similar intermediate states.

    \item \textbf{Bottle\_Handover}: The robot transfers a bottle from one side of the workspace to the other. This task evaluates whether the policy can maintain stable execution across a continuous handover-like manipulation process.

    \item \textbf{Food\_Replacement}: The robot removes the original food item, grasps a replacement item, and places it onto the plate. This task requires recognizing the transition between removal and replacement stages.

    \item \textbf{Cover\_and\_Reveal}: The robot first covers a block with a cover and then removes the cover to reveal it. This task creates strong state ambiguity because the same object may become temporarily hidden and must be recovered through temporal context.
\end{itemize}

\subsection{Additional Quantitative and Qualitative Results}
\label{app:experiment}

\paragraph{Robustness Analysis under Randomized Settings}
Table~\ref{tab:hard_results} provides the task-level performance breakdown that underpins the significant average gains reported in the main text.
As expected, the introduction of severe environmental randomization leads to a general performance decline across all methods compared to the clean setting.
However, SeedPolicy demonstrates superior robustness relative to the baselines.
While the baselines struggle with visual domain shifts, the SEG mechanism acts as an adaptive filter, effectively suppressing irrelevant visual disturbances to preserve the integrity of the latent state.
For example, in tasks like \textit{Grab Roller} and \textit{Handover Mic}, SeedPolicy retains a success rate of 51\% and 23\% respectively, whereas the Transformer baseline drops to near zero.
Even with the CNN backbone, we observe improved resilience in tasks such as \textit{Shake Bottle} (23\% vs. 8\%).
These results indicate that the dynamic gating capability of SEG is crucial for distinguishing semantic task features from environmental noise, thereby maintaining policy functionality better than standard stacking approaches.

\begin{table*}[t]
\centering
\caption{\textbf{Per-task performance comparison between Baseline and Ours on 50 tasks in the "hard" setting.} Tasks are sorted by episode time (\textbf{Ep.Time}) (average duration of expert demonstrations in seconds). Higher success rates are highlighted in \textbf{bold}.}
\label{tab:hard_results}
\renewcommand{\arraystretch}{1.3}
\setlength{\tabcolsep}{4pt}
\footnotesize
\resizebox{\linewidth}{!}{%
\begin{tabular}{l |c c |c c| c | l | c c| c c| c}
\toprule
\multirow{2}{*}{\textbf{Task}} & \multicolumn{2}{c|}{\textbf{Transformer}} & \multicolumn{2}{c|}{\textbf{CNN}} & \multirow{2}{*}{\textbf{Ep.Time}} & 
\multirow{2}{*}{\textbf{Task}} & \multicolumn{2}{c|}{\textbf{Transformer}} & \multicolumn{2}{c|}{\textbf{CNN}} & \multirow{2}{*}{\textbf{Ep.Time}} \\

\cmidrule(lr){2-3} \cmidrule(lr){4-5} \cmidrule(lr){8-9} \cmidrule(lr){10-11}

 & \textbf{Baseline} & \textbf{Ours} & \textbf{Baseline} & \textbf{Ours} & & 
 & \textbf{Baseline} & \textbf{Ours} & \textbf{Baseline} & \textbf{Ours} & \\
\midrule

Click Alarmclock          & 4 & \textbf{6} & 5 & \textbf{10} & 2 & Place Mouse Pad           & 0 & 0 & 0 & 0 & 5 \\
Click Bell                & 1 & 1 & 0 & \textbf{3} & 2 & Place Shoe                & 0 & 0 & 0 & 0 & 5 \\
Beat Block Hammer         & 0 & \textbf{9} & 0 & 0 & 3 & Rotate Qrcode             & 1 & 1 & 0 & 0 & 5 \\
Grab Roller               & 0 & \textbf{51} & 0 & 0 & 3 & Scan Object               & 0 & 0 & 0 & 0 & 5 \\
Lift Pot                  & 1 & \textbf{5} & 0 & \textbf{1} & 3 & Open Laptop               & 2 & \textbf{3} & 0 & \textbf{1} & 6 \\
Move Playingcard Away     & 1 & 1 & 0 & 0 & 3 & Handover Mic              & 0 & \textbf{23} & 0 & 0 & 7 \\
Turn Switch               & 1 & \textbf{5} & 1 & \textbf{5} & 3 & Place Bread Basket        & 0 & \textbf{2} & 0 & 0 & 7 \\
Adjust Bottle             & 9 & \textbf{17} & 0 & 0 & 4 & Place Dual Shoes          & 0 & \textbf{1} & 0 & 0 & 7 \\
Move Pillbottle Pad       & 0 & 0 & 0 & 0 & 4 & Place Burger Fries        & 0 & \textbf{8} & 0 & 0 & 8 \\
Pick Diverse Bottles      & 0 & \textbf{2} & 0 & \textbf{2} & 4 & Place Can Basket          & 0 & \textbf{3} & 0 & 0 & 8 \\
Pick Dual Bottles         & 0 & \textbf{3} & 0 & \textbf{1} & 4 & Place Object Basket       & 0 & 0 & 0 & 0 & 8 \\
Place Object Scale        & 0 & 0 & 0 & 0 & 4 & Shake Bottle              & 12 & \textbf{22} & 8 & \textbf{23} & 8 \\
Place Object Stand        & 0 & 0 & 0 & 0 & 4 & Handover Block            & 0 & \textbf{1} & 0 & 0 & 9 \\
Place Phone Stand         & 0 & 0 & 0 & 0 & 4 & Place Cans Plasticbox     & 0 & 0 & 0 & 0 & 9 \\
Press Stapler             & 13 & \textbf{26} & 0 & \textbf{2} & 4 & Put Object Cabinet        & 0 & \textbf{5} & 0 & 0 & 9 \\
Stamp Seal                & 0 & 0 & 0 & 0 & 4 & Shake Bottle Horizontally & \textbf{25} & 10 & 18 & \textbf{22} & 9 \\
Dump Bin Bigbin           & 1 & \textbf{2} & 0 & 0 & 5 & Stack Blocks Two          & 0 & \textbf{1} & 0 & 0 & 10 \\
Move Can Pot              & 0 & 0 & 0 & \textbf{2} & 5 & Stack Bowls Two           & 0 & \textbf{1} & 0 & 0 & 10 \\
Move Stapler Pad          & 0 & 0 & 0 & 0 & 5 & Hanging Mug               & 0 & 0 & 0 & 0 & 11 \\
Place A2B Left            & 0 & 0 & 0 & 0 & 5 & Open Microwave            & 0 & 0 & 0 & \textbf{3} & 14 \\
Place A2B Right           & 0 & 0 & 0 & \textbf{1} & 5 & Blocks Ranking RGB        & 0 & 0 & 0 & 0 & 15 \\
Place Bread Skillet       & 1 & 1 & 0 & 0 & 5 & Blocks Ranking Size       & 0 & 0 & 0 & 0 & 15 \\
Place Container Plate     & 0 & 0 & 0 & 0 & 5 & Stack Blocks Three        & 0 & \textbf{1} & 0 & 0 & 15 \\
Place Empty Cup           & 0 & 0 & 0 & 0 & 5 & Stack Bowls Three         & 0 & \textbf{2} & 0 & 0 & 15 \\
Place Fan                 & 0 & 0 & 0 & 0 & 5 & Put Bottles Dustbin       & 0 & \textbf{1} & 0 & \textbf{1} & 20 \\

\bottomrule
\end{tabular}
}
\end{table*}

\begin{figure*} 
    \centering
    \includegraphics[width= 1\linewidth]{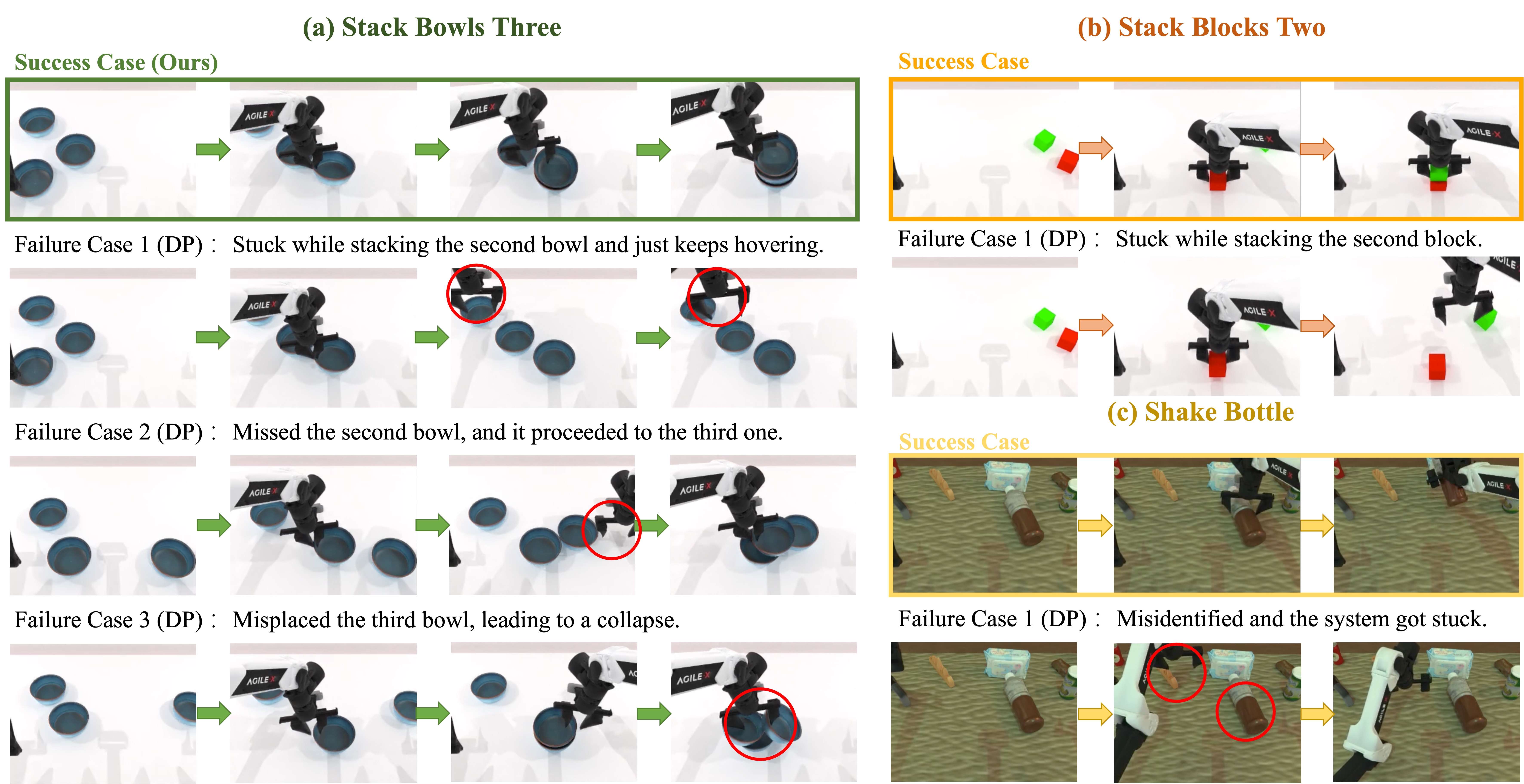}
    \caption{\textbf{More qualitative visualization of failure cases in simulation.} We compare the successful execution of SeedPolicy (top row) against representative failure modes of the DP across three tasks: (a) \textit{Stack Bowls Three} ("clean" setting), (b) \textit{Stack Blocks Two} ("clean" setting), and (c) \textit{Shake Bottle} ("hard" setting). Red circles highlight critical errors, including execution stagnation (getting stuck) and spatial positioning failures (collisions or air grabs).}
    \label{fig:appendix simulation failcase}
\end{figure*}

\begin{figure*} 
    \centering
    \includegraphics[width= 1\linewidth]{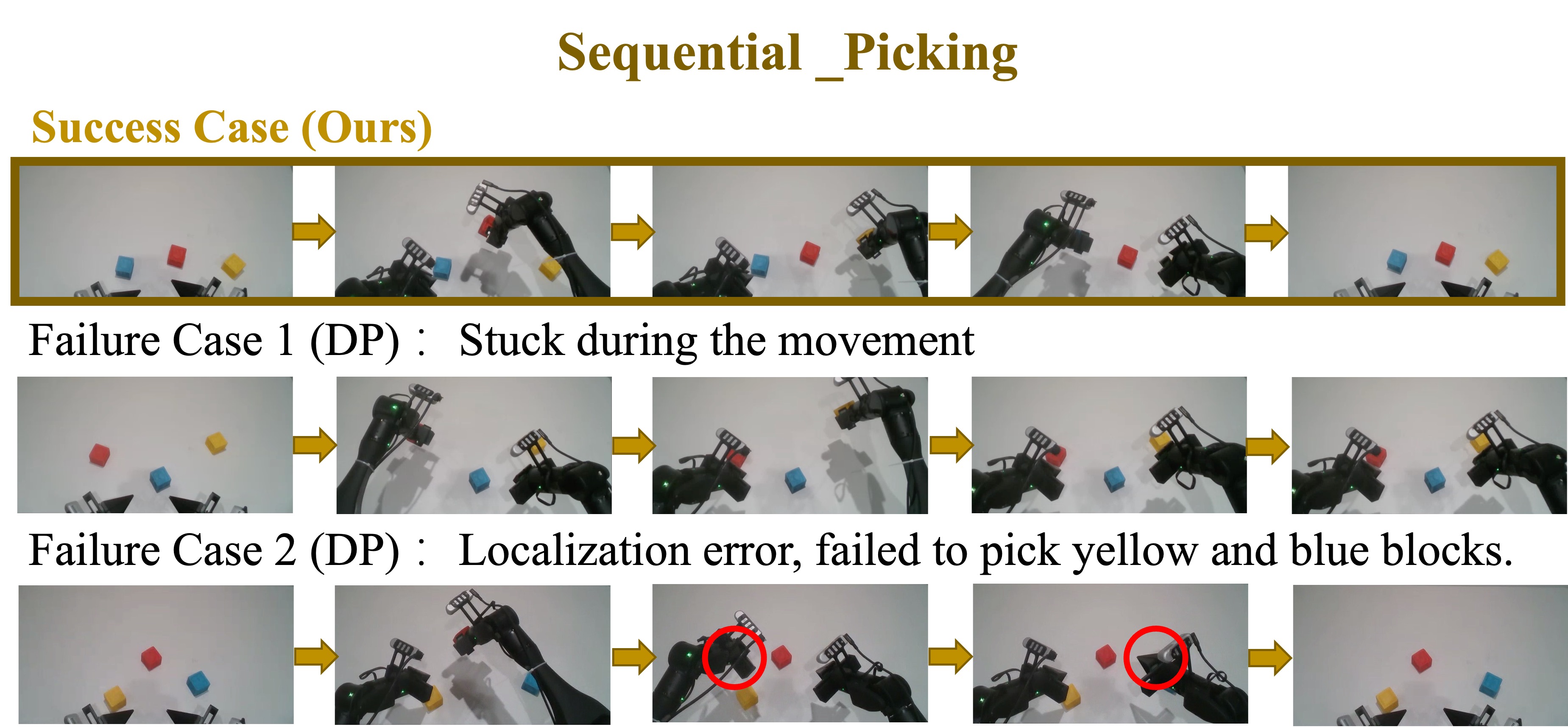}
    \caption{\textbf{More qualitative failure analysis in real-world scenarios.} We visualize the successful execution of SeedPolicy (top rows) compared to common baseline failures in \textit{Sequential\_Picking}. Red circles highlight critical errors. 
    \textbf{Failure Case 1} illustrates execution stagnation caused by \textit{Perceptual Aliasing} (misinterpreting the placed block as the initial state). \textbf{Failure Case 2} demonstrates spatial precision errors (air grabs) attributed to the lack of explicit depth information.}
    \label{fig:appendix real world failcase}
\end{figure*}

\paragraph{More qualitative failure analysis}
Supplementing the analysis in the main text, we provide additional visualizations of representative failure cases in Fig.~\ref{fig:appendix simulation failcase} and Fig.~\ref{fig:appendix real world failcase}.

In simulation tasks such as \textit{Stack Bowls Three} (Fig.~\ref{fig:appendix simulation failcase}) (a), the baseline frequently exhibits \textbf{execution stagnation}, hovering indefinitely due to its inability to track task progression over long horizons.

This limitation is further evidenced in the real-world \textit{Sequential\_Picking} task (Fig.~\ref{fig:appendix real world failcase}), where \textbf{perceptual aliasing} causes the baseline to misinterpret an intermediate state as the initial state, leading to a deadlock (Failure Case 1).

Additionally, we observe consistent \textbf{spatial precision errors} (e.g., air grabs in Failure Case 2), confirming that without the temporal depth inference provided by SeedPolicy, standard 2D baselines struggle to resolve spatial ambiguities in both simulated and physical environments.

\paragraph{Robustness under Hard-trained/Hard-tested conditions.}

In the main manuscript, the Hard setting is mainly used to evaluate zero-shot robustness, where policies trained on clean demonstrations are tested under randomized environments. To further examine whether the proposed mechanism remains effective when the training data also contains environmental randomization, we additionally train and evaluate the models under the Hard setting. As shown in Table~\ref{tab:hard_setting_rebuttal}, SeedPolicy consistently outperforms both DP and the state-based variant with an FFN gate across all representative tasks. The improvement is especially clear on long-horizon tasks such as \textit{Stack Bowls Two}, where the vanilla DP fails completely while SeedPolicy achieves a substantially higher success rate. These results further verify that the proposed cross-attention-based gating mechanism improves robustness under challenging randomized conditions.

\begin{table}[t]
\centering
\setlength{\tabcolsep}{4pt}
\renewcommand{\arraystretch}{0.9}
\caption{Success rates (\%) of models trained and evaluated in the ``Hard'' setting.}
\label{tab:hard_setting_rebuttal}
\small
\begin{tabular}{lccc}
\toprule
Task & DP & DP+State & SeedPolicy \\
     &    & +FFN gate & (Ours) \\
\midrule
Turn Switch      & 12 & 30 & \textbf{35} \\
Place Empty Cup & 0  & 3  & \textbf{6}  \\
Stack Bowls Two & 0  & 18 & \textbf{25} \\
\bottomrule
\end{tabular}
\end{table}

\paragraph{Scalability with more expert demonstrations.}

We further evaluate the data scalability of SeedPolicy by increasing the number of expert demonstrations from 50 to 100 and 200 under the manuscript setting. As reported in Table~\ref{tab:data_scaling_rebuttal}, SeedPolicy consistently benefits from more demonstrations. The gains are particularly significant for medium- and long-horizon tasks, where richer demonstrations provide more diverse temporal trajectories for learning state evolution. For example, the success rate of \textit{Place Empty Cup} increases from 32\% to 87\%, and \textit{Stack Bowls Two} improves from 73\% to 94\%. The short-horizon task \textit{Turn Switch} shows a smaller improvement because its performance is already close to saturation with 50 demonstrations. These results indicate that SeedPolicy can effectively leverage additional expert data and exhibits favorable scalability.

\begin{table}[t]
\centering
\setlength{\tabcolsep}{6pt}
\renewcommand{\arraystretch}{0.9}
\caption{Success rates (\%) of SeedPolicy under varying numbers of expert demonstrations.}
\label{tab:data_scaling_rebuttal}
\small
\begin{tabular}{lccc}
\toprule
Task & 50 Demos & 100 Demos & 200 Demos \\
\midrule
Turn Switch      & 54 & 55 & 57 \\
Place Empty Cup & 32 & 67 & 87 \\
Stack Bowls Two & 73 & 91 & 94 \\
\bottomrule
\end{tabular}
\end{table}

\paragraph{Horizon Extrapolation on Minute-level Tasks.}

To further evaluate whether SeedPolicy can handle genuinely long-horizon manipulation, we conduct additional experiments on three minute-level tasks from RoboTwin 2.0~\cite{15} and RMBench~\cite{48}. 
Different from the second-level tasks in the main benchmark, these tasks require successful executions lasting over one minute, involving substantially longer temporal dependencies and more extended action sequences. 
As shown in Table~\ref{tab:minute_level}, SeedPolicy consistently outperforms DP, ACT and $\pi_{0.5}$ on all three tasks, demonstrating stronger robustness in minute-level manipulation scenarios.

More importantly, these evaluations also test horizon extrapolation beyond the training context length. 
During training, the maximum accumulated recurrent context corresponds to an observation horizon of 120 frames. 
However, successful test rollouts can be much longer than this training horizon. 
For example, on \textit{Battery Try}, SeedPolicy achieves successful rollouts of about 64 seconds, corresponding to approximately 640 frames, which is far beyond the 120-frame maximum context used during training. 
This result indicates that SeedPolicy does not merely memorize a fixed training horizon, but can continuously update and utilize its evolving latent state during longer test-time executions.

\begin{table}[t]
\centering
\captionsetup{font=small}
\caption{Success rate (\%) comparison on three minute-level manipulation tasks.}
\label{tab:minute_level}
\small
\setlength{\tabcolsep}{4pt}
\renewcommand{\arraystretch}{0.9}
\begin{tabular}{lccccc}
\toprule
Task & DP & ACT & $\pi_{0.5}$ & SeedPolicy \\
\midrule
Swap Blocks          & 11 & 2 & 24 & \textbf{26} \\
Cover Blocks         & 0  & 0 & 0 & \textbf{15} \\
Battery Try          & 9  & 19 & 16 & \textbf{33}  \\
\bottomrule
\end{tabular}
\end{table}

\paragraph{Cross-Benchmark Generalization on MimicGen.}

To examine whether the effectiveness of SeedPolicy generalizes beyond a single simulation benchmark, we conduct additional experiments on MimicGen~\cite{50} following the setting of EquiDiff~\cite{49}. 
Different from RoboTwin 2.0, this benchmark is built in MuJoCo with a single-arm Franka robot and includes distinct manipulation tasks. 
We evaluate three representative tasks and further test two observation modalities, including image-based observations and voxel-based observations. 
As shown in Table~\ref{tab:mimicgen_tasks}, SeedPolicy consistently outperforms DP and ACT across all tasks. 
These results suggest that the proposed recurrent latent-state modeling is not tied to a specific benchmark, robot embodiment, or observation modality.

\begin{table}
\centering
\captionsetup{font=small}
\caption{Success rate (\%) comparison on MimicGen. Img and Voxel denote image-based and voxel-based observation modalities, respectively.}
\label{tab:mimicgen_tasks}
\small
\setlength{\tabcolsep}{4pt}
\renewcommand{\arraystretch}{0.9}
\begin{tabular}{lcccc}
\toprule
Task & DP & ACT & Ours-Img & Ours-Voxel \\
\midrule
Coffee D2      & 26 & 33 & \textbf{80} & 79 \\
Kitchen D1     & 50 & 61 & 83 & \textbf{89} \\
Mug Cleanup D1 & 23 & 31 & 76 & \textbf{77} \\
\bottomrule
\end{tabular}
\end{table}

\paragraph{Compatibility with Vision-Language-Action Models}

To examine whether the proposed recurrent latent-state mechanism can extend beyond diffusion-policy backbones, we integrate it into a VLA-based policy and obtain \textbf{SeedVLA}. 
Specifically, SeedVLA replaces the original memory mechanism in MemoryVLA~\cite{46} with our recurrent latent-state mechanism, while keeping the training setting unchanged. 
Each task is trained with 50 clean expert demonstrations on RoboTwin 2.0. 
As shown in Table~\ref{tab:vla_compatibility}, SeedVLA achieves the best overall performance across the five representative tasks. 
Compared with the memory-based VLA (MemoryVLA~\cite{46}), the improvement indicates that the proposed recurrent latent-state mechanism provides a more effective way to organize and update temporal context. 
Meanwhile, compared with VLA baselines without explicit memory mechanisms, such as RDT~\cite{22} and $\pi_{0.5}$~\cite{51}, SeedVLA shows that incorporating temporal memory can further enhance VLA policies. 
These results provide direct evidence that our mechanism is not limited to DP-style policies, but can also serve as a temporal module for VLA-based policies.

\begin{table*}[t]
\centering
\caption{Compatibility study on RoboTwin 2.0 with VLA-based policies.}
\label{tab:vla_compatibility}
\small
\setlength{\tabcolsep}{4pt}
\renewcommand{\arraystretch}{0.9}
\begin{tabular}{lcccc}
\toprule
Task & RDT~\cite{22} & MemoryVLA~\cite{46} & $\pi_{0.5}$~\cite{51} & SeedVLA \\
\midrule
Turn Switch         & 35 & 40 & 42 & \textbf{50} \\
Handover Mic         & 90 & 69 & 58 & \textbf{91} \\
Put Object Cabinet  & 33 & 38 & 54 & \textbf{60} \\
Stack Bowls Two     & 76 & 81 & 93 & \textbf{94} \\
Blocks Ranking Size & 0 & 27 & 36 & \textbf{45} \\
\bottomrule
\end{tabular}
\end{table*}

\subsection{More Details and Results}

\paragraph{Initialization and Resetting of the Recurrent Latent State.}
At model construction time, $S_0$ is initialized once from a zero-mean Gaussian distribution with standard deviation $0.02$ under a fixed random seed, and is optimized jointly with the rest of the policy parameters during training.

During both training and inference, the recurrent state is reset to this learned initial latent state at the beginning of each episode. 
In mini-batch training, we use the episode reset indicator to replace the carried-over state with $S_0$ whenever a new episode starts; otherwise, the state is propagated recurrently across consecutive steps. 
At inference time, the policy starts from $S_0$ at the first control step and then updates the latent state online through SEGA as new observation features arrive. 
Therefore, the initial latent state is deterministic after model initialization and is not randomly re-sampled for each episode.

\paragraph{Statistical significance analysis.}
To further examine whether the performance gains of SeedPolicy are consistent
across tasks rather than being driven by a small number of outliers, we conduct
a paired non-parametric sign test on the per-task success rates in the RoboTwin
2.0 clean setting. For each of the 50 tasks, we compare SeedPolicy with the
corresponding DP baseline under the same backbone and count the number of
tasks where SeedPolicy obtains a higher, lower, or equal success rate. Ties are
excluded when computing the two-sided exact sign-test p-value.

As shown in Table~\ref{tab:sign_test}, SeedPolicy achieves statistically
significant improvements over DP for both backbones. With the Transformer
backbone, SeedPolicy obtains 38 wins, 5 losses, and 7 ties over DP, yielding a
two-sided exact sign-test p-value of $2.5 \times 10^{-7}$. With the CNN
backbone, SeedPolicy obtains 41 wins, 6 losses, and 3 ties, yielding a p-value
of $1.8 \times 10^{-7}$. These results indicate that the improvements are
consistent across the 50 tasks and are unlikely to be explained by a few
favorable tasks.

\begin{table}[t]
\centering
\caption{Statistical significance analysis on RoboTwin 2.0 clean-setting tasks.
We perform a paired two-sided exact sign test over per-task success rates between
SeedPolicy and the corresponding DP baseline. Ties are excluded when computing
the p-value.}
\label{tab:sign_test}
\begin{tabular}{lcccc}
\toprule
Backbone & Wins & Losses & Ties & Two-sided $p$-value \\
\midrule
Transformer & 38 & 5 & 7 & $2.5 \times 10^{-7}$ \\
CNN         & 41 & 6 & 3 & $1.8 \times 10^{-7}$ \\
\bottomrule
\end{tabular}
\end{table}

\paragraph{Open-Loop Trajectory Reconstruction.} 

We further conduct open-loop evaluations on the training set to verify the model's capacity. 

\begin{figure*}[t]
    \centering
    \begin{minipage}{0.52\linewidth}
        \centering
        \includegraphics[width=\linewidth]{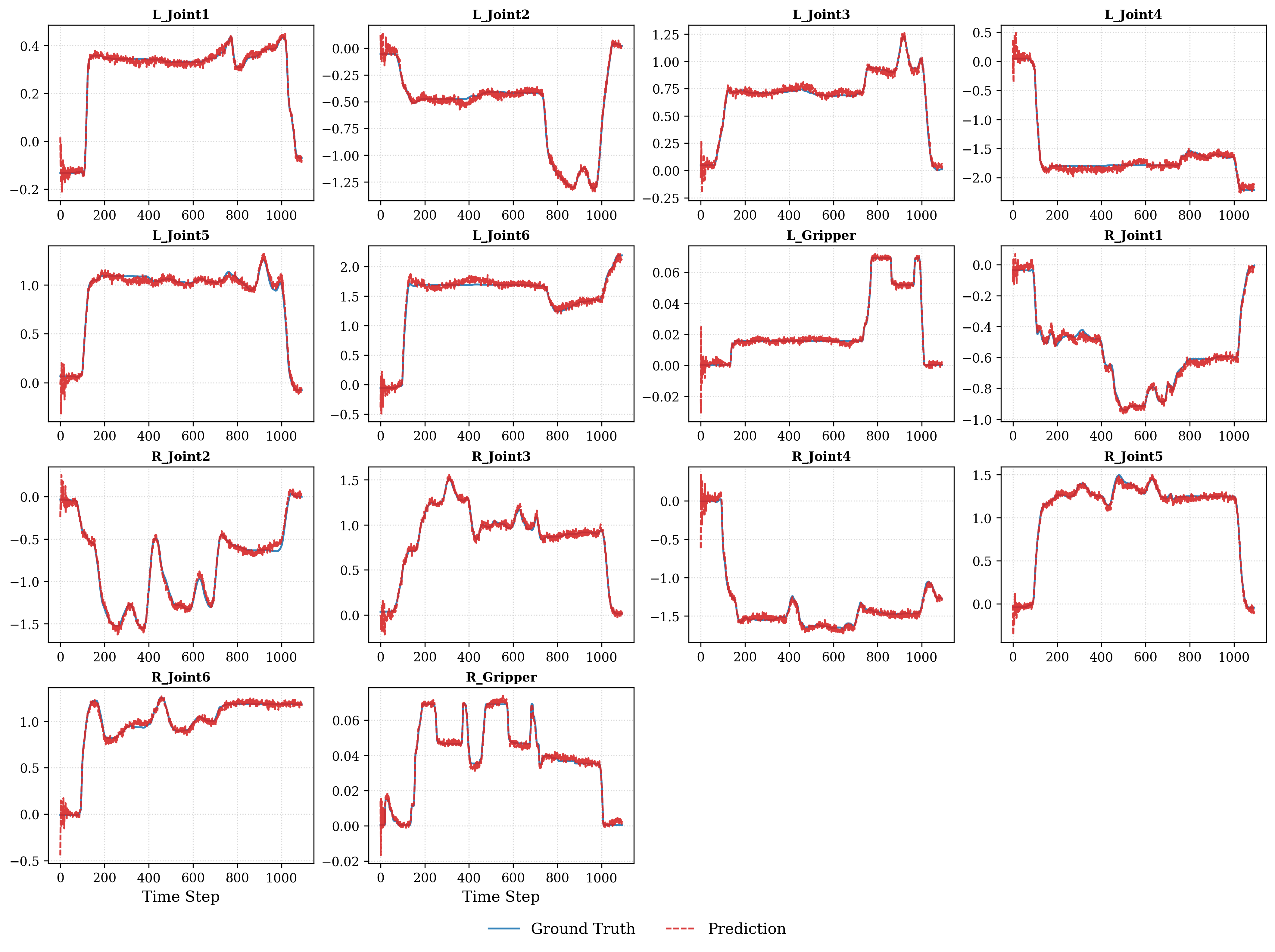}
        \caption{\textbf{Open-loop trajectory reconstruction for Sequential\_Picking.} The model accurately reconstructs the complex over 1000-step trajectory.}
        \label{fig:openloop_sequential}
    \end{minipage}
    \hfill
    \begin{minipage}{0.52\linewidth}
        \centering
        \includegraphics[width=\linewidth]{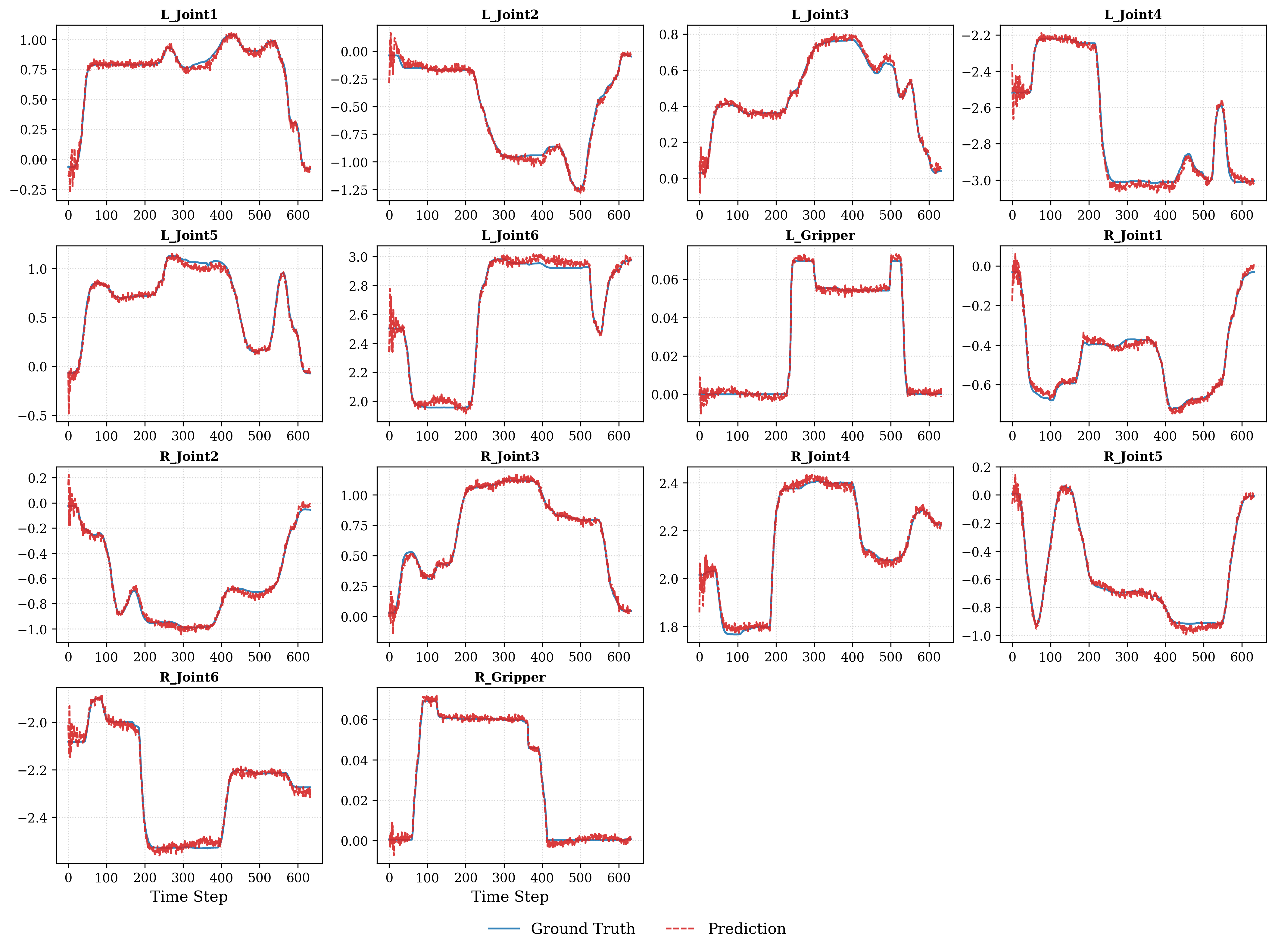}
        \caption{\textbf{Open-loop trajectory reconstruction for Bottle\_Handover.} Note the precise alignment in gripper channels, demonstrating the model's ability to capture sharp discrete transitions.}
        \label{fig:openloop_handover}
    \end{minipage}
    \hfill
    \begin{minipage}{0.52\linewidth}
        \centering
        \includegraphics[width=\linewidth]{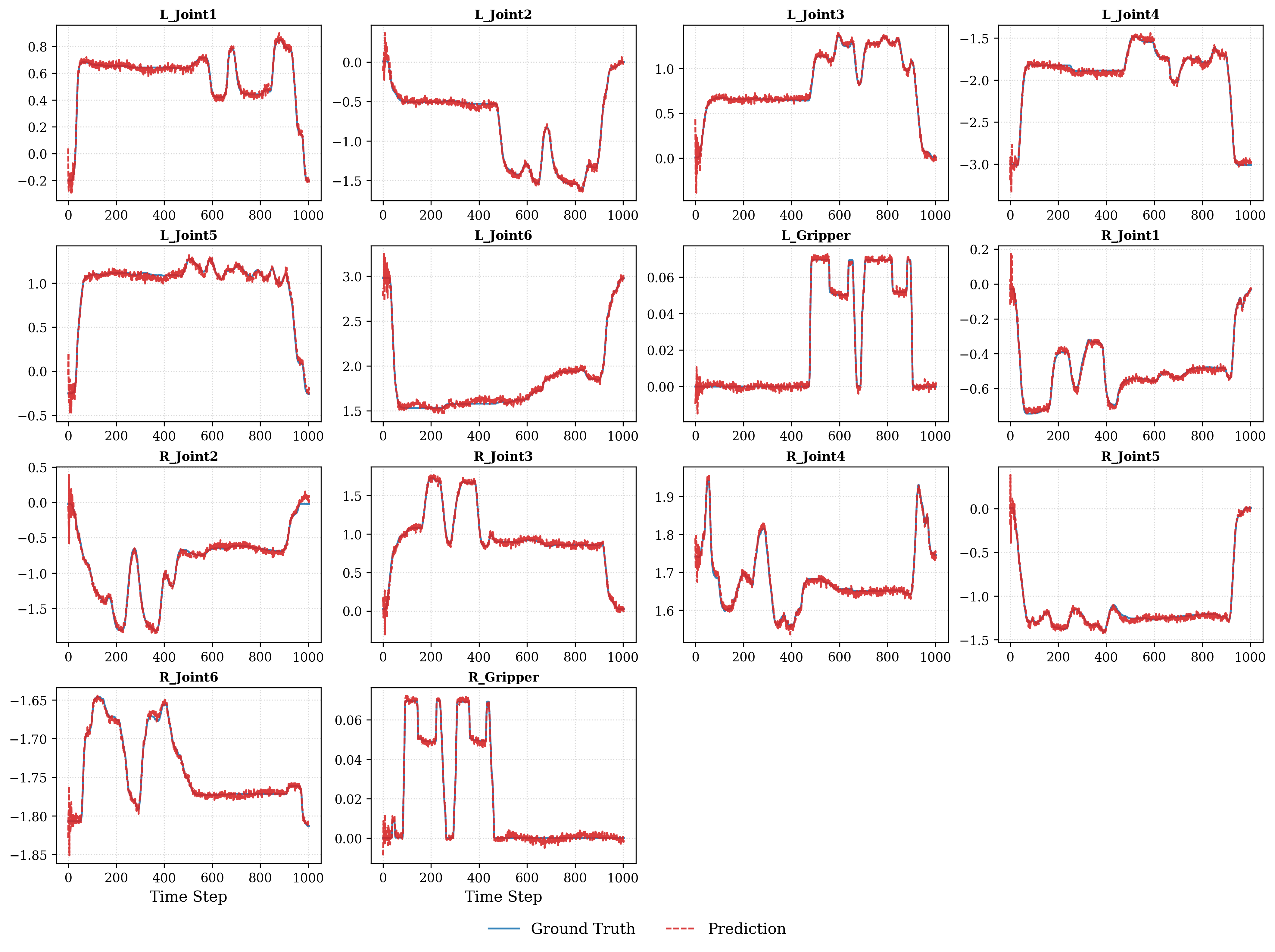}
        \caption{\textbf{Open-loop trajectory reconstruction for Looping\_Place-Retrieval.} SeedPolicy maintains high tracking accuracy over long horizons.}
        \label{fig:openloop_looping}
    \end{minipage}
    \label{fig:openloop_results}
\end{figure*}

As shown in Fig.~\ref{fig:openloop_sequential}, Fig.~\ref{fig:openloop_handover} and Fig.~\ref{fig:openloop_looping}, the predicted action trajectories (red dashed lines) exhibit high-fidelity alignment with the ground truth expert actions (blue solid lines) across all three challenging tasks. 

Even in long-horizon scenarios spanning over 1,000 steps (e.g., \textit{Sequential\_Picking}), the model maintains precise tracking without drift. 

Additionally, it accurately reconstructs sharp state transitions in gripper dimensions, indicating that SeedPolicy effectively captures the complex multi-modal distributions of expert behaviors without underfitting.

\end{document}